\pdfoutput=1

\documentclass[11pt]{article}

\usepackage[]{acl}
\usepackage{graphicx}
\usepackage{times}
\usepackage{latexsym}

\usepackage[T1]{fontenc}
\usepackage{hyperref}       
\usepackage{url}            
\usepackage{booktabs}       
\usepackage{amsmath}
\usepackage{amsfonts}       
\usepackage{nicefrac}       
\usepackage{algorithm}
\usepackage{algpseudocode}
\usepackage{wrapfig}
\usepackage{verbatim}
\usepackage{fancyvrb}
\usepackage{multicol}
\usepackage{listings}
\usepackage{microtype}      
\usepackage{xcolor}         
\usepackage{enumitem}
\usepackage{xspace}
\usepackage{arydshln}
\newif\ifshowcomment
     \showcommenttrue
\newcommand{\benchmark}{CURE\includegraphics[width=0.02\textwidth]{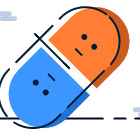}\xspace}
\ifshowcomment
    \newcommand{\heng}[1]{\textcolor{red}{[heng: #1]}}
    \newcommand{\yang}[1]{\textcolor{blue}{[yang: #1]}}
    \newcommand{\mac}[1]{\textcolor{orange}{[michael: #1]}}
    \newcommand{\ksikka}[1]{\textcolor{red}{[ksikka: #1]}}
    \newcommand{\todo}[1]{\textcolor{red}{[TODO: #1]}}
    \newcommand{\yy}[1]{\textcolor{red}{[#1]}}
    
\else
    \newcommand{\yang}[1]{}
    \newcommand{\focus}[1]{}
    \newcommand{\mac}[1]{}
    \newcommand{\ksikka}[1]{}
    \newcommand{\heng}[1]{}
    \newcommand{\todo}[1]{}
    \newcommand{\yy}[1]{}
\fi

\usepackage[utf8]{inputenc}

\usepackage{microtype}

\usepackage{inconsolata}

\title{Measuring and Improving Chain-of-Thought
Reasoning \\ in Vision-Language Models}

\author{%
 Yangyi Chen$^{1,2}$\thanks{\xspace \xspace Work done during internship at SRI International.}, 
 Karan Sikka$^{1}$, 
 Michael Cogswell$^{1}$, 
 Heng Ji$^{2}$,
 Ajay Divakaran$^{1}$\\
  $^{1}$ SRI International
  $^{2}$ University of Illinois Urbana-Champaign\\
{\tt yangyic3@illinois.edu} 
}

\begin{document}
\maketitle
\begin{abstract}
%

Vision-language models (VLMs) can effectively act as visual assistants, interpreting questions about images and producing human-like responses. This work explores their abilities to demonstrate human-like reasoning. 
To address concerns about the consistency of VLMs' reasoning, we introduce a chain-of-thought (CoT) consistency measure.
%
We tackle the challenge of extensive human annotations by proposing an LLM-Human-in-the-Loop pipeline.
%
Based on this pipeline, we build the \textbf{\benchmark} benchmark to measure both the zero-shot reasoning performance and consistency of VLMs. We evaluate state-of-the-art VLMs and find that even the best-performing model is unable to demonstrate strong visual reasoning capabilities and consistency, indicating that substantial efforts are required to enable VLMs to perform visual reasoning as systematically and consistently as humans. As an early step, we propose a two-stage training framework aimed at improving both the reasoning performance and consistency of VLMs without human annotations. 
The framework consists of two primary stages: supervised fine-tuning and learning from feedback, to guide VLMs in generating reasoning chains that exhibit both consistency and groundedness.
Our framework exhibits a 4\% relative improvement in reasoning performance and consistency.
We release the dataset at \url{https://github.com/Yangyi-Chen/CoTConsistency}.


%

%

%
%
%

\end{abstract}
\section{Introduction}

%
%
\looseness=-1
Vision-language models (VLMs) exhibit competence at generating human-like responses by leveraging multimodal instructional data and large language models (LLMs)~\cite{DBLP:journals/corr/abs-2301-12597, DBLP:journals/corr/abs-2304-08485, liu2023aligning, chen2023dress}. 
%
%
A key direction in improving such VLMs is to enable grounded and consistent visual reasoning.  
%
We thus take a critical look at the reasoning capability of existing VLMs, measuring and improving both their performance and consistency in reasoning. 
%
For reasoning performance, we aim to measure whether VLMs can derive high-level inference that extends beyond the immediately perceived information correctly. 
For reasoning consistency, we seek to determine the extent to which VLMs can identify the underlying reasoning chains that lead to the high-level inference. 





%
Previous work simplifies the evaluation of reasoning consistency by only considering coarse-grained rationales~\cite{zellers2019recognition} and relying on human evaluation~\cite{DBLP:conf/nips/LuMX0CZTCK22} and similarity measure~\cite{DBLP:journals/corr/abs-2307-12626}, which lacks scalability and preciseness.
Thus, we motivate to establish a new benchmark dataset that provides annotation of the fine-grained reasoning steps to automatically measure reasoning consistency. 
%
However, collecting such a dataset is challenging due to high-cost underlying human effort and may contain inconsistencies among annotators for the reasoning chains~\cite{DBLP:journals/ijon/GonzalezOA21, DBLP:conf/coling/LarsonCMLK20}.
%




%

%

%

 \begin{figure*}[t!]
\centering
\includegraphics[width=\textwidth]{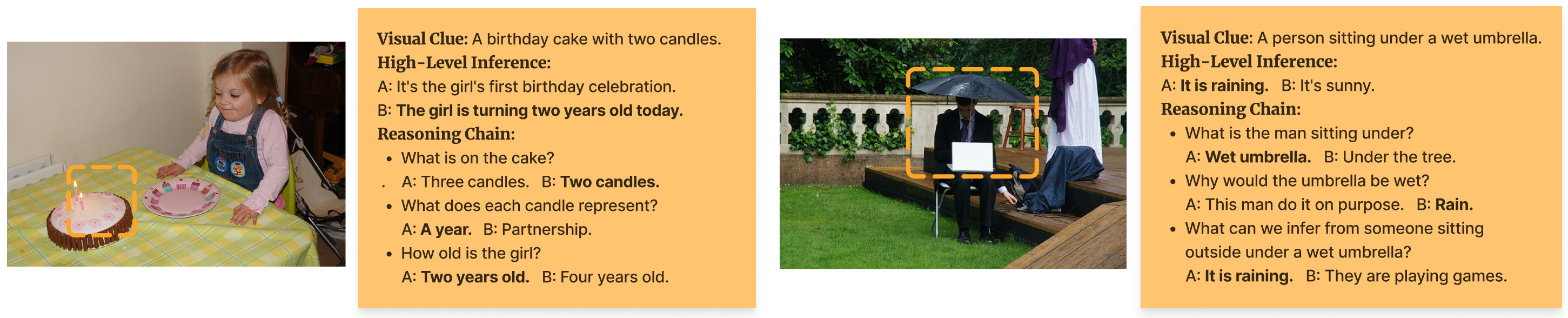}
 \caption{\looseness=-1 Besides the high-level inference about the images (e.g., \textit{The girl is turning two years old today.}), \benchmark also contains CoT reasoning chains to evaluate VLMs' reasoning performance and consistency. We only show 2 (of 6) candidate options for presentation. We highlight the ground truth answers. More examples are shown in Figure~\ref{fig:more_examples}.
} 
 \vspace{-15pt}
 \label{fig:examples}
 \end{figure*}
 
To address this challenge, we propose an LLM-Human-in-the-Loop
pipeline for dataset construction. Several recent efforts have shown that LLMs can effectively follow human instructions to generate high-quality datasets~\cite{DBLP:conf/nips/BrownMRSKDNSSAA20, DBLP:conf/nips/MengHZH22, DBLP:journals/corr/abs-2304-14334, DBLP:journals/corr/abs-2212-10560}.
This pipeline functions by incorporating limited
human assistance for providing instructions and filtering rules, enabling LLMs to efficiently generate
high-quality datasets in a semi-automatic manner, substantially reducing annotation cost.
%
Based on an existing coarse-grained visual inference dataset Sherlock~\cite{DBLP:conf/eccv/HesselHPZBRSC22}, we establish a benchmark \benchmark for \textbf{C}hain-of-Thought Vis\textbf{U}Al \textbf{R}easoning \textbf{E}valuation. 
\looseness=-1
It contains 1,622 human-verified samples of high-level visual inference and corresponding CoT reasoning chains, intended for zero-shot evaluation. 
%
Two examples are presented in Figure~\ref{fig:examples}.
Particularly, the CoT reasoning chains consist of progressive subquestions, ranging from recognition (e.g., \textit{What is on the cake?}) to cognition (e.g., \textit{What does each candle represent?}), with the purpose of measuring the reasoning consistency of VLMs.
%
%
%
Due to the notorious difficulty of natural language generation evaluation~\cite{DBLP:journals/csur/SaiMK23, DBLP:conf/iclr/HendrycksBBZMSS21}, we formulate \benchmark as a multiple-choice task for the ease of automatic evaluation. 
Particularly, for each visual input, we assess the reasoning in VLMs by evaluating their overall inference capabilities for a designated area (the bounding box in Figure~\ref{fig:examples}) and their ability to correctly address the intermediate reasoning chain leading to the final inference.


\looseness=-1
We evaluate the state-of-the-art (SOTA) VLMs on \benchmark. 
The key conclusions from these evaluations are:
(1) The model's success in complex visual inference depends on LLMs components, visual inputs, and instruction finetuning;
(2) Even the SOTA VLM (BLIP-2) falls short in comparison to human performance regarding overall visual reasoning performance.
In addition, our findings indicate a lack of reasoning consistency. Specifically, the reliability
of intermediate reasoning steps cannot be assured, irrespective of the accuracy of the final inference (and
vice versa).
This suggests VLMs are not always consistent in their reasoning.


To enhance VLMs' reasoning performance and consistency, we propose a two-stage training framework for training rationale-augmented VLMs.
In the first stage, VLMs are trained on reasoning samples that encompass step-by-step reasoning chains, which are automatically generated by LLMs.
However, VLMs may produce inaccurate high-level inferences due to inconsistencies or hallucination in the rationales after this stage. 
%
%
Thus, we introduce a subsequent stage that integrates feedback from LLMs to examine the reasoning process.
This approach avoids the complex task of directly scrutinizing the high-level inferences of VLMs.
%
The results demonstrate 
%
the relative improvement in both reasoning performance and consistency is approximately 4\% compared to the SOTA.
\section{Related Work}

The CoT reasoning approach is first proposed for LLMs~\citep{DBLP:conf/nips/Wei0SBIXCLZ22}.
We discuss related work regarding LLMs CoT reasoning and vision-language pretraining in Appendix~\ref{sec:related} and focus on vision-language reasoning in this section.
There exists a paucity of comprehensive diagnostic studies concerning VLMs with the aim of quantifying their reasoning consistency, although efforts have been spent on measuring the visual reasoning performance (e.g., Sherlock)~\cite{DBLP:conf/eccv/HesselHPZBRSC22} and coarse-grained rationale evaluation, including multiple-choice question answering (e.g., VCR)~\cite{zellers2019recognition}, human evaluation of generated rationales~\cite{DBLP:conf/nips/LuMX0CZTCK22}, and similarity measure between the generated and the ground-truth rationales~\cite{DBLP:journals/corr/abs-2307-12626}.
Some work has identified the failure of VLMs to accurately answer subquestions that are components of the main problems~\cite{DBLP:conf/emnlp/RaySDLB19, DBLP:conf/cvpr/JingJWLW22, DBLP:conf/cvpr/SelvarajuTPHRNK20, DBLP:conf/emnlp/WangYHLCC22, DBLP:conf/nips/LuMX0CZTCK22, DBLP:journals/corr/abs-2307-12626}.
For instance, VLMs may correctly determine the significant size of a mountain in an image but erroneously classify it as small when responding to a query such as "Are the mountains small?"~\cite{DBLP:conf/emnlp/RaySDLB19}. 
In contrast to the aforementioned studies that focus on coarse-grained rationale evaluation and individual subquestions, we create reasoning chains that consist of coherent subquestions capable of supporting high-level inference. This approach allows us to precisely measure the extent to which reasoning in VLMs is consistent and grounded.

 \begin{figure*}[t!]
\centering
\includegraphics[width=0.95\textwidth]{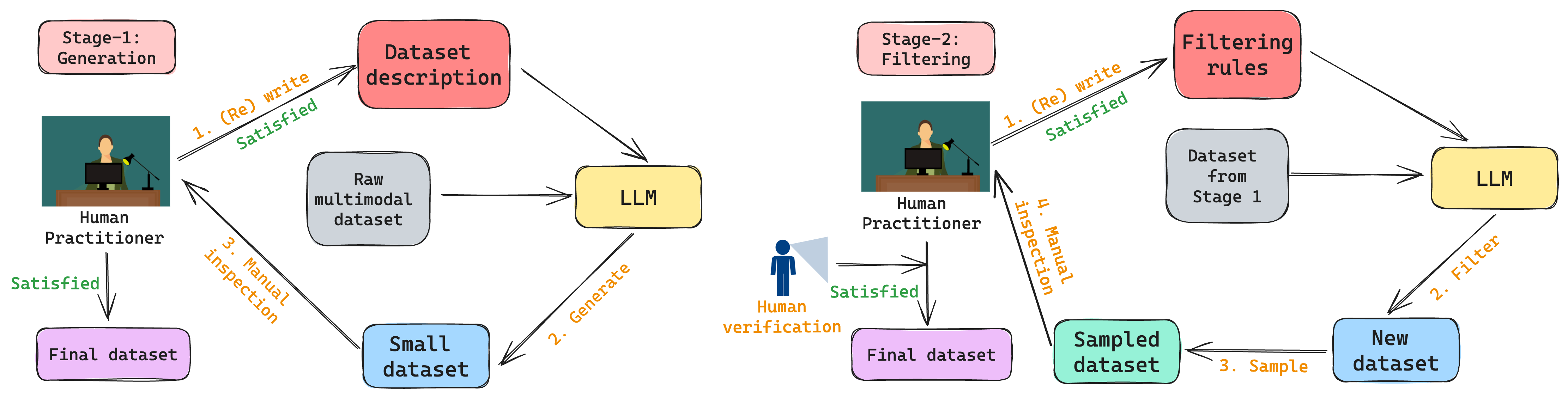}
 \caption{
 The LLM-Human-in-the-Loop dataset construction pipeline consists of the generation and filtering stages. We use this procedure to create \benchmark in a semi-automatic manner.
 } 
 \label{fig:llm-human-in-the-loop}
 \end{figure*}




\section{\benchmark \  Benchmark}
We present the \benchmark dataset for measuring visual reasoning performance and consistency in VLMs and the LLM-Human-in-the-Loop pipeline adopted to construct it semi-automatically.
Our dataset builds on the Sherlock dataset~\cite{DBLP:conf/eccv/HesselHPZBRSC22}, which measures abductive reasoning by annotating visual clues (text and bounding boxes for perceptual elements) and high-level inference.
%
However, our aim is not only to measure the capacity of VLMs to accurately perform high-level visual inference but also to subsequently ascertain the extent to which the resulting inference is thoroughly substantiated.
We thus add two new annotations to enable this:
\textbf{(1) Reasoning Chains:} 
We provide fine-grained and precise CoT reasoning containing coherent subquestions that can be chained together to derive the high-level inference provided by Sherlock. 
%
\textbf{(2) Candidate Answers:} 
To avoid the long-standing issues in the evaluation of natural language generation~\cite{DBLP:journals/csur/SaiMK23}, we transform the generation task of high-level inference and CoT subquestions into a multiple-choice question answering task by generating plausible but incorrect alternative candidates for each ground truth, as shown in Figure~\ref{fig:examples}.

\looseness=-1
In this section, we outline the procedure to semi-automatically create \benchmark with LLMs and then describe the evaluation metrics adopted to measure reasoning performance and consistency.

\subsection{LLM-Human-in-the-Loop Data Generation Pipeline}
\label{sec:llm_human_intheloop}
Our dataset construction pipeline consists of two stages, as illustrated in Figure~\ref{fig:llm-human-in-the-loop}.
The first stage aims to generate a preliminary dataset that potentially contains instances of failure, while the second stage filters out the error cases, similar to the crowdsourcing dataset collection approaches~\cite{DBLP:conf/eccv/LinMBHPRDZ14}.
In both stages, LLMs carry out the majority of tasks, while human practitioners (the researchers in this case) iteratively correct errors made by LLMs~\cite{DBLP:journals/corr/abs-2303-12712}.
%
%
%
%


%

%
%

%

\subsubsection{Stage-1: Preliminary Generation}
We randomly select 10,000 examples from the Sherlock evaluation set to serve as the raw coarse-grained examples. 
In this stage, the practitioner engineers an initial prompt that basically describes the data LLMs should generate based on each raw example.
The dataset description is then fed along with necessary context -- the visual clues describing the image and the high-level inference from Sherlock -- to generate a small initial dataset of reasoning chains (e.g., for 50 examples).
These examples are usually inadequate and look different than intended.
%
%
Next, the practitioner should carefully examine the generated examples and revise the dataset description accordingly.
Through multiple iterations, a curated instruction that contains dataset descriptions and specific requirements can be produced to guide LLMs to generate the full-sized preliminary dataset.


\paragraph{Reasoning Chains.} 
We use GPT-4~\cite{DBLP:journals/corr/abs-2303-08774} in all dataset generation steps.
Our stage-1 prompt for generating reasoning steps is shown in Appendix~\ref{sec:prompt}. 
This prompt starts by describing the overall goal, inputs, and outputs we expect from LLMs.
It then outlines five principles to ensure LLMs generate meaningful and reasonable subquestions. 
We also find that the inclusion of an in-context example for a step-by-step demonstration of sample generation significantly enhances the ability of LLMs to generate samples that conform to the specified principles.
%
The resulting preliminary dataset contains fairly uniform reasoning chains for 1.6k examples.
Typically the generated subquestions support the high-level inference when chained together, following a progression from perception problems to more complex visual inference, thus adhering to the "from recognition to cognition" practice~\cite{zellers2019recognition}.


\begin{table*}[tbp!]
\centering
\resizebox{0.9\textwidth}{!}{
\begin{tabular}{cl}
\toprule
Iteration & \multicolumn{1}{c}{Common Failure Modes}                                                                \\ \midrule
1         & The CoT reasoning chains lack consistent subquestions that are capable of deriving the high-level inference. \\
2         & The candidate inference about the image exhibits similarity in meaning with the ground truth inference.      \\
3         & The ground truth answers for the subquestions are incorrect due to the occurrence of hallucination in LLMs.              \\
4         & The candidate answers for the subquestions are also correct.                    \\
5         & The problems can be solved directly without relying on visual inputs.                                          \\
6         & The subquestions can contain some words that are irrelevant to the visual inputs.                        \\ \bottomrule
\end{tabular}
}
\caption{The identified common failure modes at each iteration.}
\vspace{-10pt}
\label{tab:filter}
\end{table*}
\paragraph{Candidate Answers.}
We can potentially evaluate whether the outputs from VLMs match or closely resemble ground truth inference or reasoning steps, similar to the practice in previous work~\cite{DBLP:conf/nips/LuMX0CZTCK22, DBLP:journals/corr/abs-2307-12626}.
However, this approach has two notable shortcomings:
(1) The evaluation of natural language generation has been a persistent challenge, lacking a universally accepted approach~\cite{DBLP:journals/csur/SaiMK23}; 
(2) Although we provide ground truth answers for each image, some alternative predictions may also be correct, regarding the nature of abductive reasoning~\cite{walton2014abductive}. 
To address the above issues, we formulate \benchmark as a multiple-choice question answering task, requiring VLMs to select the most likely inference/answer from the six candidates provided.  
We prompt LLMs using the same stage-1 procedure to generate potential candidate inference/answers.
%
These candidate answers maintain relevance to the provided image while incorporating factual inaccuracies when compared to the ground truth. 
The prompts adopted are shown in Appendix~\ref{sec:prompt}.
%
%

\subsubsection{Stage-2: Filtering}
\looseness=-1
Although samples in the preliminary dataset generally adhere to the desired criteria, failures still arise due to inherent limitations in LLMs~\cite{DBLP:journals/corr/abs-2302-03494}.
However, by drawing explicit attention to common failure modes, we can instruct LLMs to correctly filter out bad example groups.
In each round, the practitioner selects a small number of samples and conducts a thorough inspection to extract predominant failure modes.
A distinct prompt is then created for each failure mode that requires LLMs to determine whether reasoning chains or sets of candidate answers meet that failure case.
%
This prompt is applied to all remaining preliminary data, removing all examples that LLMs identify as lying in the failure modes. 
%
The practitioner then repeats this procedure through multiple iterations until the randomly selected sample of examples no longer exhibits any instances of error.
We conduct a total of six iterations to systematically remove groups of samples that displayed common failure modes.
The identified failure modes are listed in Table~\ref{tab:filter}, and the prompts are described in Appendix~\ref{sec:prompt}.

%



\paragraph{Human Verification.}
While the filtering stage yields a substantial labor reduction when compared to the initial unfiltered dataset (50\% reduction estimated), there still exist some failure cases.
For example, our analysis finds that a certain amount of examples in the Sherlock dataset share the same reasoning problem that relies on simplistic visual cues such as sky and lighting conditions to infer weather patterns and differentiate between day and night.
This kind of shortcut annotation is documented in previous studies~\cite{DBLP:conf/naacl/GururanganSLSBS18, DBLP:conf/emnlp/GevaGB19, DBLP:conf/acl/YuanZC023}.
We motivate to address these concerns since \benchmark is for evaluation purposes. 
We hire human annotators to meticulously review the entire created dataset to ensure two primary objectives: 
(1) Each sample's validity for measuring reasoning performance and consistency; 
(2) The inclusion of diverse samples in the evaluation dataset.
The details of human verification are described in Appendix~\ref{sec:human_annotation}.
%

\subsection{Human Evaluation}
\benchmark contains 1,622 evaluative instances. We employ human annotators to conduct human evaluation with emphasis on two aspects:
(1) What is the level of human performance observed on \benchmark?
(2) Do the samples within \benchmark hold validity and can be effectively used for evaluation?
We select a sample of 200 instances from \benchmark.
The annotation details are described in Appendix~\ref{sec:human_annotation}.
We engage three human annotators to conduct the task of answering multiple-choice questions and provide annotations indicating the presence of any failure mode mentioned in Table~\ref{tab:filter} or any other unidentified failure modes.
The human performance is listed in Table~\ref{tab:exp}.
The detailed discussion of the human performance compared with the model performance is in Sec.~\ref{sec:exp}. 
In the assessment of sample validity, merely 3\% of the evaluation samples within the benchmark are found to demonstrate specific issues. Of this subset, 2\% of the samples exhibit inconsistent reasoning chains, while 1\% of the samples provide incorrect answers for the subquestions.
It is worth noting that apart from the issues outlined in Table~\ref{tab:filter}, no other problems have been reported.
These findings serve as a validation of the high quality of \benchmark, and also demonstrate the effectiveness of our pipeline at identifying unqualified samples. 
The detailed statistics of \benchmark are described in Appendix~\ref{sec:stat}.

\subsection{Evaluation Metrics}
\label{sec:eval_metrics}
\looseness=-1
As described in the previous section, we frame \benchmark as a multiple-choice problem with six potential inference per image and six plausible candidates for every subquestion (reasoning step). 
Specifically, each image $I_i$ is paired with a high level question $Q_h^i$ associated with six candidate inferences \(O_h^i = \{o_{h1}^i, o_{h2}^i, ..., o_{h6}^i\}\). 
Additionally, reasoning chains are made up of several questions $Q_c^i$. Each question \(q \in Q_c^i\) is associated with a set of six candidate answers \(O_q^i = \{o_{q1}^i, o_{q2}^i, ..., o_{q6}^i\}\). 
We propose a series of metrics that evaluate not only the reasoning ability of the VLMs but also the consistency in their reasoning.

%
\subsubsection{Metrics for Reasoning Performance}
\textbf{Performance in High-Level Reasoning.} 
The metric $R_h$ is designed to measure the VLMs' ability in accurately choosing the most probable inference from the candidate pool for each image: 
\begin{equation}
\begin{split}
R_h &= \frac{1}{N}\sum_{i=1}^{N}\mathbb{I}(\hat{a}_h^i = a_h^i), \\ \hat{a}_h^i &\in \{o_{h1}^i, o_{h2}^i, ..., o_{h6}^i\},
\end{split}
\end{equation}
where $N$ signifies the total number of images, $\mathbb{I}(x)$ is the indication function that returns 1 if x is true and 0 otherwise, \(\hat{a}_h^i\) and \(a_h^i\) are model's chosen answer and ground truth answer respectively.


\textbf{Performance in CoT Reasoning.} The metric $R_{cot}$ is used to evaluate the VLMs' ability to correctly answer all subquestions contained in the reasoning chain for each image: 
\begin{equation}
\begin{split}
R_{\text{cot}} &= \frac{1}{N}\sum_{i=1}^{N}\mathbb{I}\left(\sum_{j=1}^{M} \mathbb{I}(\hat{a}_j^i = a_{j}^i) = M\right), \\
\hat{a}_j^i &\in \{o_{j1}^i, o_{j2}^i, \ldots, o_{j6}^i\},
\end{split}
\end{equation}
where \(M\) is the number of subquestions within the CoT reasoning chain per image, $\hat{a}_j^i$ is the model's prediction, and $a_{j}^i$ is the ground truth answer.

\textbf{Overall Performance in Reasoning.} 
We propose $R_o$ to measure if VLMs can successfully perform both high-level reasoning and CoT reasoning for every image: 
\begin{equation}
R_{o} = \frac{1}{N}\sum_{i=1}^{N}\mathbb{I}(\hat{a}_h^i = a_h^i) * \mathbb{I}(\sum_{j=1}^{M} \mathbb{I}(\hat{a}_j^i = a_{j}^i) = M)
\end{equation}
where the notations adhere to previous definitions. 

\subsubsection{Metrics for Reasoning Consistency}
\textbf{Consistency in Forward Reasoning.}
We define $C_f$ to evaluate the VLMs' capacity to correctly answer the high-level inference question once all subquestions have been correctly addressed: 
\begin{equation}
\begin{split}
C_f &= \frac{1}{\sum_{i=1}^{N} s_i} \sum_{i=1}^{N} s_i \cdot \mathbb{I}(\hat{a}_h^i = a_{h}^i), \\ \hat{a}_h^i &\in \{o_{q1}^i, o_{q2}^i, ..., o_{q6}^i\},
\end{split}
\end{equation}
where \(s_i\) equals 1 if all subquestions for the \(i\)th image have been correctly answered by the VLM, and 0 otherwise, and other notations adhere to their previous definitions.

\textbf{Consistency in Backward Reasoning.}
We define \(C_b\) to evaluate the VLMs' proficiency in correctly answering all subquestions given the successful answering of the high-level inference question: 
\begin{equation}
\begin{split}
C_b &= \sum_{i=1}^{N} \mathbb{I}\left(\sum_{j=1}^{M} \mathbb{I}(\hat{a}_j^i = a_{j}^i) = M\right), \\
\hat{a}_j^i &\in \{o_{j1}^i, o_{j2}^i, \ldots, o_{j6}^i\},
\end{split}
\end{equation}
where \(h_i\) equals 1 if the VLM correctly answers the high-level inference question for the \(i\)th image, and 0 otherwise, and other notations adhere to their previous definitions. 

 \begin{figure*}[tbp!]
\centering
\includegraphics[width=\textwidth]{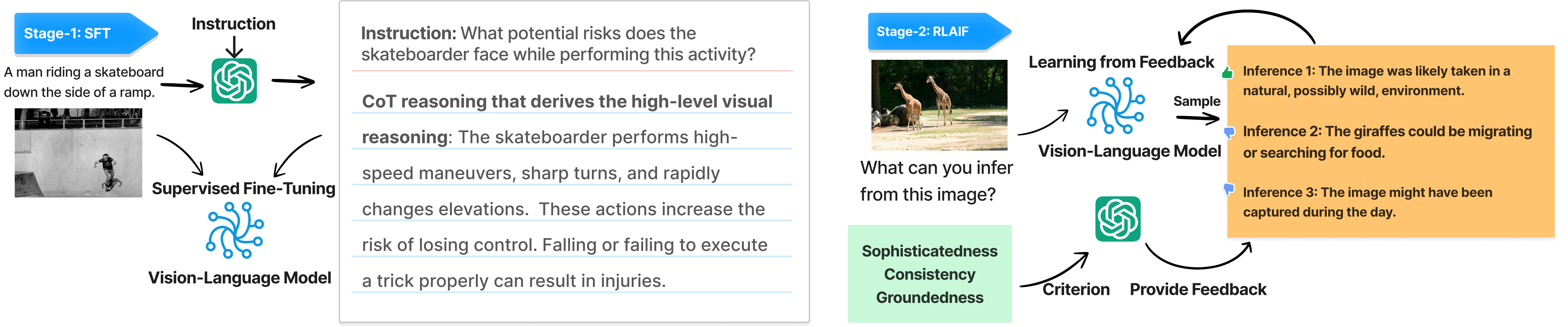}
 \caption{
The two-stage training framework consisting of SFT and RLAIF. 
We instruct LLMs to examine the reasoning process of VLMs to improve the challenging high-level inferences.
 } 
 \label{fig:approach}
 \end{figure*}

\section{Approach}
\label{sec:approach}
\looseness=-1
In preliminary experiments, we find that VLMs can effectively conduct high-level visual inference when provided with complete reasoning chains.
Thus, we propose to train a model capable of generating rationales that can potentially enhance visual reasoning performance and consistency. 
%
To this end, we propose a bifurcated training framework that is able to train a VLM to efficiently produce rationales that facilitate high-level visual inference (see Figure~\ref{fig:approach}).
In the first stage, we aim to train CoTBLIP to generate rationales that contain enough visual details and reasonable inference. 
%
To further mitigate certain issues in generated rationales (e.g., hallucination) for better high-level inferences, we introduce the second Reinforcement Learning from LLMs (AI) Feedback (\textbf{RLAIF}) stage. 
%
We select BLIP-2-T5$_{xl}$ as our backbone model due to its strong performance on basic vision-language tasks~\cite{DBLP:journals/corr/abs-2306-09265, DBLP:journals/corr/abs-2306-13394}.
Consequently, we refer to our rationale-generation model as \textbf{CoTBLIP}.

%




\looseness=-1
\paragraph{Stage-1: SFT.}
We utilize the complex reasoning samples from the LLaVA dataset~\cite{DBLP:journals/corr/abs-2304-08485}. 
The original 77K samples are produced by instructing GPT-4 to generate visual inference using a carefully curated set of five human-annotated captions and bounding boxes associated with images from the COCO Dataset~\cite{DBLP:conf/eccv/LinMBHPRDZ14}. 
However, the generated samples consist of repetitive, dialogic expressions that might not be entirely grounded in the images. 
We thus perform a further post-processing step that prompts LLMs to generate CoT reasoning chains based on the original samples, placing an emphasis on ensuring that these chains are logical, consistent, and succinct.
The detailed prompt is shown in Appendix~\ref{sec:prompt}.
%
We train CoTBLIP on these refined samples using SFT.

Following the SFT stage, CoTBLIP is competent in generating plausible rationales.
%
However, the high-level inferences may be inaccurate since the produced rationales might contain inconsistent reasoning chains or contents that are not grounded in the images (hallucination). 
In addition, the scalability of the SFT training stage is limited due to its dependence on high-quality human-annotated dense captions, which makes it difficult for this stage to leverage image-caption pairs in the wild. This can lead to lower generalizability on a broad range of visual concepts. 
%
Therefore, we extend the training into a second stage, optimizing the generation of rationales using feedback from LLMs.
Specifically, we leverage LLMs to inspect the reasoning process, which is more straightforward than directly scrutinizing the high-level inferences.

\paragraph{Stage-2: RLAIF.}
\looseness=-1
In this stage, we use image-caption pairs sourced from the wild (e.g., SBU Captions~\cite{DBLP:conf/nips/OrdonezKB11}). 
For each image, CoTBLIP is initially prompted to generate three CoT reasoning chains, leading to high-level visual inference regarding each image. 
We also note that there is a noticeable variation in the quality of these generated reasoning chains, which necessitates external feedback. 
Therefore, we use LLMs (GPT-3.5-Turbo) to provide feedback on the reasoning chains based on the provided caption, considering three aspects:
\begin{itemize} [topsep=1pt, partopsep=1pt, leftmargin=12pt, itemsep=-1pt]
\item \textbf{Sophistication:} 
The CoT reasoning chains should derive interesting high-level visual inference, instead of trivial visual information (e.g., The image might be captured during the day.)

\item \textbf{Consistency:} The reasoning chains should be logically consistent to derive the high-level inference without unsupported assertions or gaps.

\item \textbf{Groundedness:} The extracted visual details in the reasoning chains should be fully grounded in the images, instead of hallucination.
\end{itemize}
%
The prompt we use is described in Appendix~\ref{sec:prompt}.
%
We adapt the methods proposed by~\cite{DBLP:conf/nips/Ouyang0JAWMZASR22} to facilitate pairwise comparison between two reasoning chains and establish a ranking for the three generated reasoning chains.
In addition, we leverage a consistency check to exclude instances in which LLMs exhibit conflicting rankings. 
%
We use the SBU Captions to generate around 27K LLM preference samples considering the constraints of our available resources. 
We also demonstrate that increasing the sample size during this stage results in consistent performance improvements in Section~\ref{sec:further_analysis}.

\looseness=-1
Given the LLM preference data, we employ Conditional Reinforcement Learning to train CoTBLIP due to its stability as observed in previous work~\cite{lu2022quark, liu2023chain, laskin2022context}.
Specifically, we introduce two control tokens, namely <Good> and <Bad>. 
For each sample containing a set of three ranked reasoning chains, we add the <Good> control token to the highest-ranked chain and the <Bad> control tokens to the remaining two chains.
In the training time, given an appended control token, we optimize CoTBLIP to maximize the likelihood of the associated reasoning chain.
Through this approach, CoTBLIP can learn to distinguish the difference between control tokens and their respective outputs~\cite{liu2023chain}. 
We note that there is no requirement to perform training for a separate reward model, given that the LLM is capable of fulfilling that role effectively.

\paragraph{Inference.}
During inference, we initially prompt CoTBLIP to generate rationales. 
However, it is important to acknowledge that when dealing with CoT subquestions that primarily involve basic visual perceptual problems and text-only inference based on provided visual details, the generated rationales may have limited effectiveness.
Thus, the rationales are used exclusively for high-level visual inference.
Specifically, these rationales are incorporated before the top-tier question to prompt the downstream VLMs to generate the prediction. 
In our implementation, we opt for utilizing the original BLIP-2-T5$_{xl}$ model to conduct predictions based on the rationales generated by CoTBLIP.
%

%

\section{Experiment}
\label{sec:exp}

\subsection{Model}
We evaluate the reasoning performance and consistency of the following models on \benchmark. 
We include GPT-3.5-Turbo-0613 (\textbf{Turbo}), which is a text-only model without visual inputs.
We include \textbf{OFA-Large/Huge}~\cite{DBLP:conf/icml/WangYMLBLMZZY22}, which are the leading VLMs without LLMs component.
We include the \textbf{BLIP-2-OPT$_{6.7b}$/T5$_{xl}$}~\cite{DBLP:journals/corr/abs-2301-12597}, which effectively utilizes LLMs for vision-language modeling. Additionally, we incorporate \textbf{InstructBLIP-T5$_{xl}$}~\cite{DBLP:journals/corr/abs-2305-06500}, which performs instruction tuning on a mixture of vision-language datasets.
We include \textbf{LLaVA$_{13b}$}~\cite{DBLP:journals/corr/abs-2304-08485} and \textbf{miniGPT-4$_{13b}$}~\cite{DBLP:journals/corr/abs-2304-10592} that have undergone extensive training on vision-language instruction tuning data.
Our approach \textbf{CoTBLIP} appends the generated CoT reasoning chain to the frozen BLIP-2-T5$_{xl}$ model and prompts it to predict the answer.
Note that this pertains exclusively to high-level visual inference. 











%



%

\begin{figure*}
  \begin{minipage}[b]{0.54\textwidth}
    \centering
    \includegraphics[width=\textwidth]{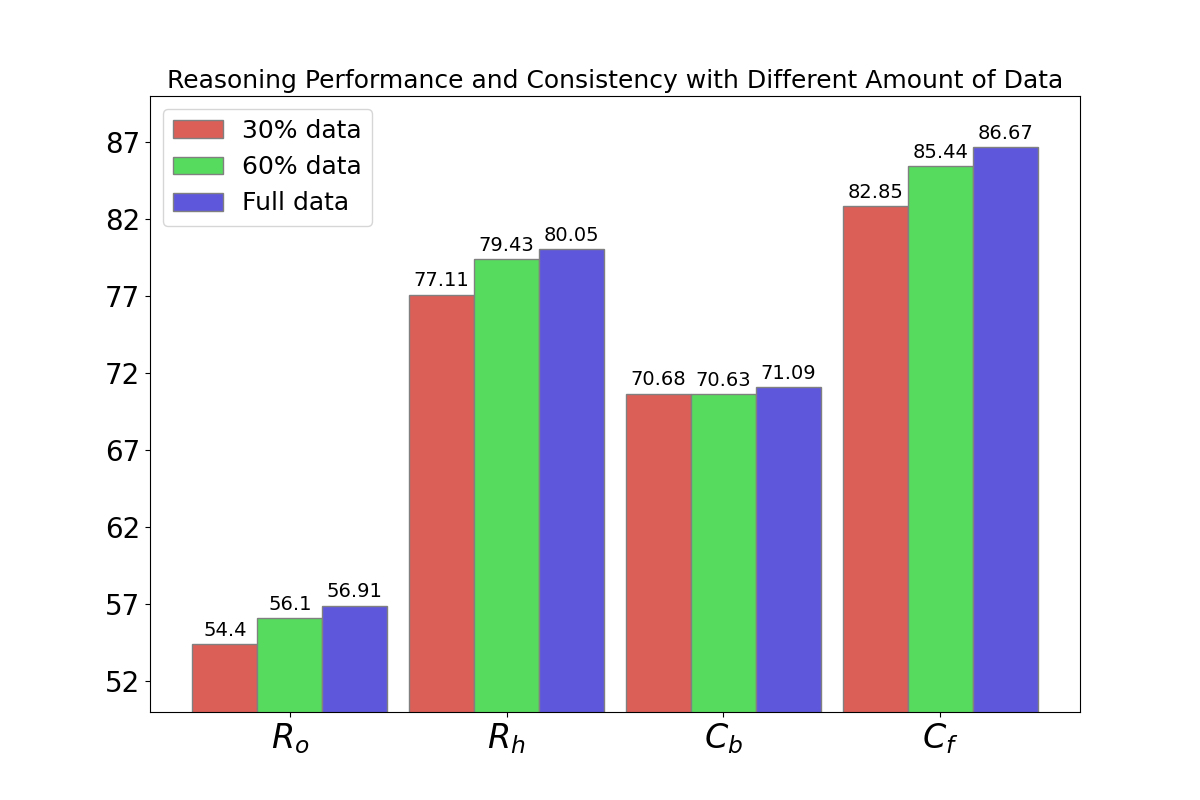} 
    \caption{The influence of the percentage of training samples in RLAIF stage on performance.}
    \label{fig:training_data}
  \end{minipage}
  \hfill
  \begin{minipage}[b]{0.45\textwidth}
    \centering
    \includegraphics[width=\textwidth]{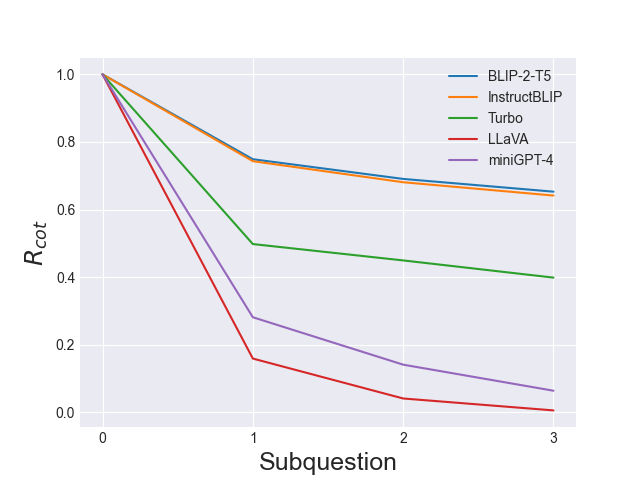} 
    \caption{The CoT reasoning performance across the subquestions. }
    \label{fig:cot_reasoning_consistency}
  \end{minipage}
\end{figure*}

\subsection{Experimental Results}
The concrete implementation details of evaluation are described in Appendix~\ref{sec:implementation}.
We consider the evaluation metrics defined in Sec.~\ref{sec:eval_metrics}.
The experimental results regarding the reasoning performance and consistency are listed in Table~\ref{tab:exp}.
We summarize the findings as follows:
(1) The model's ability to perform complex visual inference and produce reasonable outputs relies on three crucial elements: LLMs, visual inputs, and instruction fine-tuning. 
Models solely reliant on text-based information (Turbo), VLMs lacking LLMs components (OFA), and VLMs incorporating LLMs that have not undergone instructional fine-tuning (BLIP-2-OPT) exhibit inadequate performance;
(2) The Chat-based VLMs (LLaVA, miniGPT-4) that have been explicitly supervised fine-tuned on synthetic user-interaction response samples exhibit a lack of visual reasoning ability and reasoning consistency. 
The underlying cause can be ascribed to the informal nature of the chat-style data, which lacks sufficient supervision to facilitate VLMs in acquiring the ability to integrate visual elements effectively for performing high-level visual inference;
(3) The existing best-performing model, BLIP-2-T5, still falls short in reasoning performance and consistency, compared to the human evaluation results. 
This suggests that significant effort  is needed to facilitate VLMs in achieving a level of visual reasoning comparable to that of humans in a systematic and consistent manner;
(4) Our framework improves VLMs' ability to perform visual reasoning and demonstrate better reasoning consistency to a certain extent.
Specifically, we observe a 4\% improvement in both the high-level visual inference and the forward reasoning consistency. 
CoTBLIP offers a distinct advantage by providing CoT rationales that contain both extracted visual details and potential inference, thereby improving the visual reasoning pertaining to a specific image.

\begin{table}[t!]
\centering
\resizebox{0.5\textwidth}{!}{
\begin{tabular}{l|ccc|cc}
\toprule
Metric           & \multicolumn{3}{c|}{Performance} & \multicolumn{2}{c}{Consistency} \\ \midrule
Model            & $R_o$      & $R_{h}$      & $R_{cot}$      & $C_b$          & $C_f$          \\ \midrule
Random           & 0.14       & 16.67          & 0.82         & 0.82           & 16.67          \\
Turbo    & 15.97      & 33.42          & 40.26        & 47.79          & 39.66          \\
OFA-Large        & 0.12       & 17.63          & 0.62         & 0.70           & 20.0           \\
OFA-Huge         & 0.06       & 16.40          & 0.68         & 0.38           & 9.09           \\
BLIP-2-OPT       & 0.06       & 14.61          & 0.62         & 0.42           & 10.0           \\
BLIP-2-T5        & 54.56      & 76.82          & \textbf{65.66}        & 71.03          & 83.10          \\
InstructBLIP-T5  & 54.01      & 76.14          & 65.35        & 70.93          & 82.64          \\
LLaVA            & 0.12       & 14.67          & 17.82        & 17.65          & 14.29          \\
miniGPT-4        & 2.10       & 23.12          & 38.75        & 41.80          & 28.81          \\
CoTBLIP (ours) & \textbf{56.91}      & \textbf{80.05}          & \textbf{65.66}        & \textbf{71.09}        & \textbf{86.67}          \\
Human            & 85.0       & 93.0           & 89.0         & 91.40          & 95.51          \\ \bottomrule
\end{tabular}
}
\caption{The results (\%) of the reasoning performance and consistency. The human performance is averaged among 3 human annotators. See Sec.~\ref{sec:eval_metrics} for the metrics.}
\label{tab:exp}
\end{table}

\subsection{Further Analysis}
\label{sec:further_analysis}
\begin{table}[t!]
\centering

\resizebox{0.4\textwidth}{!}{
\begin{tabular}{l|cc|cc}
\toprule
Metric           & \multicolumn{2}{c|}{Performance} & \multicolumn{2}{c}{Consistency} \\ \midrule
Model            & $R_o$      & $R_{h}$            & $C_b$          & $C_f$          \\ \midrule
BLIP-2-T5       &    54.93       &   77.68                        &    70.71            &   83.66           \\
CoTBLIP &  56.91      & 80.05                & 71.09        & 86.67       \\
- w/o RLAIF     &    55.06        &   78.67              &    69.98            &    83.85           \\
- w/o SFT        &     54.75          &  77.32                    &     70.81       & 83.38        \\

\bottomrule
\end{tabular}
}
\caption{\looseness=-1 Ablation study of the SFT and RLAIF stages (\%). BLIP-2-T5 refers to prompting BLIP-2 without training to generate rationales.
The $R_{cot}$ metric (omitted here) holds the same across all methods because the generated rationales are only used for high-level visual inference. 
}
\vspace{-15pt}
\label{tab:ablation}
\end{table}
\paragraph{Ablation Study.} 
We conduct an ablation study to understand the contribution of the SFT and RLAIF stages.
The results are presented in Table~\ref{tab:ablation}. 
We observe that both of these stages contribute to the improvement in reasoning performance and consistency.
In particular, we observe further improvements when employing the RLAIF after the SFT stage. For example, the overall reasoning ($R_o$) for the combined stages (CoTBLIP) is 56.91 compared to 54.93 and 55.06 by the baseline and the SFT stage only, respectively.  
This can be attributed to the ability of RLAIF to facilitate enhanced calibration of the generated rationales, thereby augmenting their cohesiveness and substantiated nature. 
However, using only the RLAIF without the SFT stage negatively impacts performance when contrasted with the results of directly prompting BLIP-2 without training for rationale generation followed by answer prediction.
The presence of the SFT stage enables VLMs to generate reasonable rationales.
In its absence, CoTBLIP (BLIP-2) is restricted to producing only image captions or trivial rationales that do not contribute significantly to high-level inference.
Thus, without the SFT stage for initialization, the training of CoTBLIP with RLAIF is not feasible.

\looseness=-1
\paragraph{Training Data of the RLAIF Stage.}
We investigate the impact of varying the amount of training data during the RLAIF stage (see Figure~\ref{fig:training_data}).
We omit the presentation of $R_{cot}$ as they are identical.
Our findings reveal that a continuous expansion of training samples positively impacts the RLAIF training stage of CoTBLIP, regarding both reasoning performance and consistency. 
These results demonstrate the potential of utilizing web-scale image-captions data to further improve the training, attributing to the scalability of the RLAIF stage.

\looseness=-1
\paragraph{Backward Reasoning Consistency.}
We conduct a comprehensive study on the CoT reasoning performance ($R_{cot}$) of VLMs, evaluating the extent of performance degradation in answering subquestions (see Figure~\ref{fig:cot_reasoning_consistency}).
We select examples that contain three subquestions for the presentation purpose. 
We observe that existing VLMs often struggle with the initial visual perceptual problem, which involves basic visual details needed for high-level visual inference. 
However, these models can partially derive the high-level inference when provided with the extracted visual details to some degree, evidenced by the relatively small performance drop when answering the second and third questions. 
%
%
This demonstrates that high-level visual inference derived by VLMs is not entirely grounded in the visual details, leading to a low $C_b$.
We also discuss the forward reasoning consistency in Appendix~\ref{sec:forward}.

\section{Conclusion}
%
We create \benchmark using an LLM-Human-in-the-Loop pipeline and identify the deficiencies in existing VLMs for reasoning performance and consistency.
To tackle these challenges, we introduce a two-stage training framework consisting of supervised fine-tuning and learning from LLMs feedback.
Our method demonstrates improvement in VLMs' reasoning performance and consistency.

\section*{Limitation}
As shown in Table~\ref{tab:exp}, our proposed CoTBLIP still exhibits a significant gap, regarding the reasoning performance and consistency compared to the human annotators. 
This indicates substantial efforts are necessary to enable existing VLMs to perform robust visual inference like humans.  
CoTBLIP currently can only generate general visual inference about the given images, without considering the instructions. 
Future work is needed to enable CoTBLIP to perform instruction-guided reasoning chain generation that can more effectively facilitate high-level inference. 

\section*{Acknowledgement}
We thank the reviewers for their suggestions and comments. This research is based upon work supported by U.S. DARPA ECOLE Program No. HR00112390060 and  U.S. DARPA KAIROS Program No. FA8750-19-2-1004. 
The views and conclusions contained herein are those of the authors and should not be interpreted as necessarily representing the official policies, either expressed or implied, of DARPA, or the U.S. Government. The U.S. Government is authorized to reproduce and distribute reprints for governmental purposes notwithstanding any copyright annotation therein.

\bibliography{custom}

\begin{thebibliography}{74}
\expandafter\ifx\csname natexlab\endcsname\relax\def\natexlab#1{#1}\fi

\bibitem[{Borji(2023)}]{DBLP:journals/corr/abs-2302-03494}
Ali Borji. 2023.
\newblock A categorical archive of chatgpt failures.
\newblock \emph{CoRR}.

\bibitem[{Brown et~al.(2020)Brown, Mann, Ryder, Subbiah, Kaplan, Dhariwal,
  Neelakantan, Shyam, Sastry, Askell, Agarwal, Herbert{-}Voss, Krueger,
  Henighan, Child, Ramesh, Ziegler, Wu, Winter, Hesse, Chen, Sigler, Litwin,
  Gray, Chess, Clark, Berner, McCandlish, Radford, Sutskever, and
  Amodei}]{DBLP:conf/nips/BrownMRSKDNSSAA20}
Tom~B. Brown, Benjamin Mann, Nick Ryder, Melanie Subbiah, Jared Kaplan,
  Prafulla Dhariwal, Arvind Neelakantan, Pranav Shyam, Girish Sastry, Amanda
  Askell, Sandhini Agarwal, Ariel Herbert{-}Voss, Gretchen Krueger, Tom
  Henighan, Rewon Child, Aditya Ramesh, Daniel~M. Ziegler, Jeffrey Wu, Clemens
  Winter, Christopher Hesse, Mark Chen, Eric Sigler, Mateusz Litwin, Scott
  Gray, Benjamin Chess, Jack Clark, Christopher Berner, Sam McCandlish, Alec
  Radford, Ilya Sutskever, and Dario Amodei. 2020.
\newblock Language models are few-shot learners.
\newblock In \emph{Advances in Neural Information Processing Systems 33: Annual
  Conference on Neural Information Processing Systems 2020, NeurIPS 2020,
  December 6-12, 2020, virtual}.

\bibitem[{Bubeck et~al.(2023)Bubeck, Chandrasekaran, Eldan, Gehrke, Horvitz,
  Kamar, Lee, Lee, Li, Lundberg, Nori, Palangi, Ribeiro, and
  Zhang}]{DBLP:journals/corr/abs-2303-12712}
S{\'{e}}bastien Bubeck, Varun Chandrasekaran, Ronen Eldan, Johannes Gehrke,
  Eric Horvitz, Ece Kamar, Peter Lee, Yin~Tat Lee, Yuanzhi Li, Scott~M.
  Lundberg, Harsha Nori, Hamid Palangi, Marco~T{\'{u}}lio Ribeiro, and
  Yi~Zhang. 2023.
\newblock Sparks of artificial general intelligence: Early experiments with
  {GPT-4}.
\newblock \emph{CoRR}.

\bibitem[{Chen et~al.(2022)Chen, Ma, Wang, and
  Cohen}]{DBLP:journals/corr/abs-2211-12588}
Wenhu Chen, Xueguang Ma, Xinyi Wang, and William~W. Cohen. 2022.
\newblock Program of thoughts prompting: Disentangling computation from
  reasoning for numerical reasoning tasks.
\newblock \emph{CoRR}.

\bibitem[{Chen et~al.(2023)Chen, Sikka, Cogswell, Ji, and
  Divakaran}]{chen2023dress}
Yangyi Chen, Karan Sikka, Michael Cogswell, Heng Ji, and Ajay Divakaran. 2023.
\newblock Dress: Instructing large vision-language models to align and interact
  with humans via natural language feedback.
\newblock \emph{arXiv preprint arXiv:2311.10081}.

\bibitem[{Dai et~al.(2023)Dai, Li, Li, Tiong, Zhao, Wang, Li, Fung, and
  Hoi}]{DBLP:journals/corr/abs-2305-06500}
Wenliang Dai, Junnan Li, Dongxu Li, Anthony Meng~Huat Tiong, Junqi Zhao,
  Weisheng Wang, Boyang Li, Pascale Fung, and Steven C.~H. Hoi. 2023.
\newblock Instructblip: Towards general-purpose vision-language models with
  instruction tuning.
\newblock \emph{CoRR}.

\bibitem[{Dou et~al.(2022)Dou, Xu, Gan, Wang, Wang, Wang, Zhu, Zhang, Yuan,
  Peng, Liu, and Zeng}]{DBLP:conf/cvpr/DouXGWWWZZYP0022}
Zi{-}Yi Dou, Yichong Xu, Zhe Gan, Jianfeng Wang, Shuohang Wang, Lijuan Wang,
  Chenguang Zhu, Pengchuan Zhang, Lu~Yuan, Nanyun Peng, Zicheng Liu, and
  Michael Zeng. 2022.
\newblock An empirical study of training end-to-end vision-and-language
  transformers.
\newblock In \emph{{IEEE/CVF} Conference on Computer Vision and Pattern
  Recognition, {CVPR} 2022, New Orleans, LA, USA, June 18-24, 2022}. {Ieee}.

\bibitem[{Fu et~al.(2023)Fu, Chen, Shen, Qin, Zhang, Lin, Qiu, Lin, Yang,
  Zheng, Li, Sun, and Ji}]{DBLP:journals/corr/abs-2306-13394}
Chaoyou Fu, Peixian Chen, Yunhang Shen, Yulei Qin, Mengdan Zhang, Xu~Lin,
  Zhenyu Qiu, Wei Lin, Jinrui Yang, Xiawu Zheng, Ke~Li, Xing Sun, and Rongrong
  Ji. 2023.
\newblock {MME:} {A} comprehensive evaluation benchmark for multimodal large
  language models.
\newblock \emph{CoRR}.

\bibitem[{Gan et~al.(2022)Gan, Li, Li, Wang, Liu, and
  Gao}]{DBLP:journals/ftcgv/GanLLWLG22}
Zhe Gan, Linjie Li, Chunyuan Li, Lijuan Wang, Zicheng Liu, and Jianfeng Gao.
  2022.
\newblock Vision-language pre-training: Basics, recent advances, and future
  trends.
\newblock \emph{Found. Trends Comput. Graph. Vis.}

\bibitem[{Geva et~al.(2019)Geva, Goldberg, and
  Berant}]{DBLP:conf/emnlp/GevaGB19}
Mor Geva, Yoav Goldberg, and Jonathan Berant. 2019.
\newblock Are we modeling the task or the annotator? an investigation of
  annotator bias in natural language understanding datasets.
\newblock In \emph{Proceedings of the 2019 Conference on Empirical Methods in
  Natural Language Processing and the 9th International Joint Conference on
  Natural Language Processing, {EMNLP-IJCNLP} 2019, Hong Kong, China, November
  3-7, 2019}. Association for Computational Linguistics.

\bibitem[{Gonz{\'{a}}lez et~al.(2021)Gonz{\'{a}}lez, Orozco{-}Gutierrez, and
  {\'{A}}lvarez{-}Meza}]{DBLP:journals/ijon/GonzalezOA21}
Juli{\'{a}}n~Gil Gonz{\'{a}}lez, {\'{A}}lvaro{-}{\'{A}}ngel Orozco{-}Gutierrez,
  and Andr{\'{e}}s {\'{A}}lvarez{-}Meza. 2021.
\newblock Learning from multiple inconsistent and dependent annotators to
  support classification tasks.
\newblock \emph{Neurocomputing}.

\bibitem[{Gururangan et~al.(2018)Gururangan, Swayamdipta, Levy, Schwartz,
  Bowman, and Smith}]{DBLP:conf/naacl/GururanganSLSBS18}
Suchin Gururangan, Swabha Swayamdipta, Omer Levy, Roy Schwartz, Samuel~R.
  Bowman, and Noah~A. Smith. 2018.
\newblock Annotation artifacts in natural language inference data.
\newblock In \emph{Proceedings of the 2018 Conference of the North American
  Chapter of the Association for Computational Linguistics: Human Language
  Technologies, NAACL-HLT, New Orleans, Louisiana, USA, June 1-6, 2018, Volume
  2 (Short Papers)}. Association for Computational Linguistics.

\bibitem[{Hendrycks et~al.(2021)Hendrycks, Burns, Basart, Zou, Mazeika, Song,
  and Steinhardt}]{DBLP:conf/iclr/HendrycksBBZMSS21}
Dan Hendrycks, Collin Burns, Steven Basart, Andy Zou, Mantas Mazeika, Dawn
  Song, and Jacob Steinhardt. 2021.
\newblock Measuring massive multitask language understanding.
\newblock In \emph{9th International Conference on Learning Representations,
  {ICLR} 2021, Virtual Event, Austria, May 3-7, 2021}. OpenReview.net.

\bibitem[{Hessel et~al.(2022)Hessel, Hwang, Park, Zellers, Bhagavatula,
  Rohrbach, Saenko, and Choi}]{DBLP:conf/eccv/HesselHPZBRSC22}
Jack Hessel, Jena~D. Hwang, Jae~Sung Park, Rowan Zellers, Chandra Bhagavatula,
  Anna Rohrbach, Kate Saenko, and Yejin Choi. 2022.
\newblock The abduction of sherlock holmes: {A} dataset for visual abductive
  reasoning.
\newblock In \emph{Computer Vision - {ECCV} 2022 - 17th European Conference,
  Tel Aviv, Israel, October 23-27, 2022, Proceedings, Part {XXXVI}}. Springer.

\bibitem[{Huang et~al.(2021)Huang, Zeng, Huang, Liu, Fu, and
  Fu}]{DBLP:conf/cvpr/HuangZH0FF21}
Zhicheng Huang, Zhaoyang Zeng, Yupan Huang, Bei Liu, Dongmei Fu, and Jianlong
  Fu. 2021.
\newblock Seeing out of the box: End-to-end pre-training for vision-language
  representation learning.
\newblock In \emph{{IEEE} Conference on Computer Vision and Pattern
  Recognition, {CVPR} 2021, virtual, June 19-25, 2021}. Computer Vision
  Foundation / {IEEE}.

\bibitem[{Huang et~al.(2020)Huang, Zeng, Liu, Fu, and
  Fu}]{DBLP:journals/corr/abs-2004-00849}
Zhicheng Huang, Zhaoyang Zeng, Bei Liu, Dongmei Fu, and Jianlong Fu. 2020.
\newblock Pixel-bert: Aligning image pixels with text by deep multi-modal
  transformers.
\newblock \emph{CoRR}.

\bibitem[{Jia et~al.(2021)Jia, Yang, Xia, Chen, Parekh, Pham, Le, Sung, Li, and
  Duerig}]{DBLP:conf/icml/JiaYXCPPLSLD21}
Chao Jia, Yinfei Yang, Ye~Xia, Yi{-}Ting Chen, Zarana Parekh, Hieu Pham,
  Quoc~V. Le, Yun{-}Hsuan Sung, Zhen Li, and Tom Duerig. 2021.
\newblock Scaling up visual and vision-language representation learning with
  noisy text supervision.
\newblock In \emph{Proceedings of the 38th International Conference on Machine
  Learning, {ICML} 2021, 18-24 July 2021, Virtual Event}. {Pmlr}.

\bibitem[{Jin and Lu(2023)}]{DBLP:conf/acl/JinL23}
Ziqi Jin and Wei Lu. 2023.
\newblock Tab-cot: Zero-shot tabular chain of thought.
\newblock In \emph{Findings of the Association for Computational Linguistics:
  {ACL} 2023, Toronto, Canada, July 9-14, 2023}. Association for Computational
  Linguistics.

\bibitem[{Jing et~al.(2022)Jing, Jia, Wu, Liu, and
  Wu}]{DBLP:conf/cvpr/JingJWLW22}
Chenchen Jing, Yunde Jia, Yuwei Wu, Xinyu Liu, and Qi~Wu. 2022.
\newblock Maintaining reasoning consistency in compositional visual question
  answering.
\newblock In \emph{{IEEE/CVF} Conference on Computer Vision and Pattern
  Recognition, {CVPR} 2022, New Orleans, LA, USA, June 18-24, 2022}. {Ieee}.

\bibitem[{Kim et~al.(2021)Kim, Son, and Kim}]{DBLP:conf/icml/KimSK21}
Wonjae Kim, Bokyung Son, and Ildoo Kim. 2021.
\newblock Vilt: Vision-and-language transformer without convolution or region
  supervision.
\newblock In \emph{Proceedings of the 38th International Conference on Machine
  Learning, {ICML} 2021, 18-24 July 2021, Virtual Event}. {Pmlr}.

\bibitem[{Lanham et~al.(2023)Lanham, Chen, Radhakrishnan, Steiner, Denison,
  Hernandez, Li, Durmus, Hubinger, Kernion, Lukosiute, Nguyen, Cheng, Joseph,
  Schiefer, Rausch, Larson, McCandlish, Kundu, Kadavath, Yang, Henighan,
  Maxwell, Telleen{-}Lawton, Hume, Hatfield{-}Dodds, Kaplan, Brauner, Bowman,
  and Perez}]{DBLP:journals/corr/abs-2307-13702}
Tamera Lanham, Anna Chen, Ansh Radhakrishnan, Benoit Steiner, Carson Denison,
  Danny Hernandez, Dustin Li, Esin Durmus, Evan Hubinger, Jackson Kernion,
  Kamile Lukosiute, Karina Nguyen, Newton Cheng, Nicholas Joseph, Nicholas
  Schiefer, Oliver Rausch, Robin Larson, Sam McCandlish, Sandipan Kundu, Saurav
  Kadavath, Shannon Yang, Thomas Henighan, Timothy Maxwell, Timothy
  Telleen{-}Lawton, Tristan Hume, Zac Hatfield{-}Dodds, Jared Kaplan, Jan
  Brauner, Samuel~R. Bowman, and Ethan Perez. 2023.
\newblock Measuring faithfulness in chain-of-thought reasoning.
\newblock \emph{CoRR}.

\bibitem[{Larson et~al.(2020)Larson, Cheung, Mahendran, Leach, and
  Kummerfeld}]{DBLP:conf/coling/LarsonCMLK20}
Stefan Larson, Adrian Cheung, Anish Mahendran, Kevin Leach, and Jonathan~K.
  Kummerfeld. 2020.
\newblock Inconsistencies in crowdsourced slot-filling annotations: {A}
  typology and identification methods.
\newblock In \emph{Proceedings of the 28th International Conference on
  Computational Linguistics, {COLING} 2020, Barcelona, Spain (Online), December
  8-13, 2020}. International Committee on Computational Linguistics.

\bibitem[{Laskin et~al.(2022)Laskin, Wang, Oh, Parisotto, Spencer, Steigerwald,
  Strouse, Hansen, Filos, Brooks et~al.}]{laskin2022context}
Michael Laskin, Luyu Wang, Junhyuk Oh, Emilio Parisotto, Stephen Spencer,
  Richie Steigerwald, DJ~Strouse, Steven Hansen, Angelos Filos, Ethan Brooks,
  et~al. 2022.
\newblock In-context reinforcement learning with algorithm distillation.
\newblock \emph{arXiv preprint arXiv:2210.14215}.

\bibitem[{Li et~al.(2020{\natexlab{a}})Li, Duan, Fang, Gong, and
  Jiang}]{DBLP:conf/aaai/LiDFGJ20}
Gen Li, Nan Duan, Yuejian Fang, Ming Gong, and Daxin Jiang. 2020{\natexlab{a}}.
\newblock Unicoder-vl: {A} universal encoder for vision and language by
  cross-modal pre-training.
\newblock In \emph{The Thirty-Fourth {AAAI} Conference on Artificial
  Intelligence, {AAAI} 2020, The Thirty-Second Innovative Applications of
  Artificial Intelligence Conference, {IAAI} 2020, The Tenth {AAAI} Symposium
  on Educational Advances in Artificial Intelligence, {EAAI} 2020, New York,
  NY, USA, February 7-12, 2020}. {AAAI} Press.

\bibitem[{Li et~al.(2023{\natexlab{a}})Li, Li, Savarese, and
  Hoi}]{DBLP:journals/corr/abs-2301-12597}
Junnan Li, Dongxu Li, Silvio Savarese, and Steven C.~H. Hoi.
  2023{\natexlab{a}}.
\newblock {BLIP-2:} bootstrapping language-image pre-training with frozen image
  encoders and large language models.
\newblock \emph{CoRR}.

\bibitem[{Li et~al.(2021{\natexlab{a}})Li, Selvaraju, Gotmare, Joty, Xiong, and
  Hoi}]{li2021align}
Junnan Li, Ramprasaath Selvaraju, Akhilesh Gotmare, Shafiq Joty, Caiming Xiong,
  and Steven Chu~Hong Hoi. 2021{\natexlab{a}}.
\newblock Align before fuse: Vision and language representation learning with
  momentum distillation.
\newblock \emph{Advances in neural information processing systems}.

\bibitem[{Li et~al.(2023{\natexlab{b}})Li, Hessel, Yu, Ren, Chang, and
  Choi}]{DBLP:conf/acl/LiHYRC023}
Liunian~Harold Li, Jack Hessel, Youngjae Yu, Xiang Ren, Kai{-}Wei Chang, and
  Yejin Choi. 2023{\natexlab{b}}.
\newblock Symbolic chain-of-thought distillation: Small models can also "think"
  step-by-step.
\newblock In \emph{Proceedings of the 61st Annual Meeting of the Association
  for Computational Linguistics (Volume 1: Long Papers), {ACL} 2023, Toronto,
  Canada, July 9-14, 2023}. Association for Computational Linguistics.

\bibitem[{Li et~al.(2019)Li, Yatskar, Yin, Hsieh, and Chang}]{li2019visualbert}
Liunian~Harold Li, Mark Yatskar, Da~Yin, Cho-Jui Hsieh, and Kai-Wei Chang.
  2019.
\newblock Visualbert: A simple and performant baseline for vision and language.
\newblock \emph{arXiv preprint arXiv:1908.03557}.

\bibitem[{Li et~al.(2021{\natexlab{b}})Li, You, Wang, Zareian, Chang, and
  Chang}]{DBLP:conf/naacl/LiYWZCC21}
Liunian~Harold Li, Haoxuan You, Zhecan Wang, Alireza Zareian, Shih{-}Fu Chang,
  and Kai{-}Wei Chang. 2021{\natexlab{b}}.
\newblock Unsupervised vision-and-language pre-training without parallel images
  and captions.
\newblock In \emph{Proceedings of the 2021 Conference of the North American
  Chapter of the Association for Computational Linguistics: Human Language
  Technologies, {NAACL-HLT} 2021, Online, June 6-11, 2021}. Association for
  Computational Linguistics.

\bibitem[{Li et~al.(2020{\natexlab{b}})Li, Yin, Li, Zhang, Hu, Zhang, Wang, Hu,
  Dong, Wei, Choi, and Gao}]{DBLP:conf/eccv/Li0LZHZWH0WCG20}
Xiujun Li, Xi~Yin, Chunyuan Li, Pengchuan Zhang, Xiaowei Hu, Lei Zhang, Lijuan
  Wang, Houdong Hu, Li~Dong, Furu Wei, Yejin Choi, and Jianfeng Gao.
  2020{\natexlab{b}}.
\newblock Oscar: Object-semantics aligned pre-training for vision-language
  tasks.
\newblock In \emph{Computer Vision - {ECCV} 2020 - 16th European Conference,
  Glasgow, UK, August 23-28, 2020, Proceedings, Part {XXX}}. Springer.

\bibitem[{Lin et~al.(2014)Lin, Maire, Belongie, Hays, Perona, Ramanan,
  Doll{\'{a}}r, and Zitnick}]{DBLP:conf/eccv/LinMBHPRDZ14}
Tsung{-}Yi Lin, Michael Maire, Serge~J. Belongie, James Hays, Pietro Perona,
  Deva Ramanan, Piotr Doll{\'{a}}r, and C.~Lawrence Zitnick. 2014.
\newblock Microsoft {COCO:} common objects in context.
\newblock In \emph{Computer Vision - {ECCV} 2014 - 13th European Conference,
  Zurich, Switzerland, September 6-12, 2014, Proceedings, Part {V}}. Springer.

\bibitem[{Liu et~al.(2023{\natexlab{a}})Liu, Lin, Li, Wang, Yacoob, and
  Wang}]{liu2023aligning}
Fuxiao Liu, Kevin Lin, Linjie Li, Jianfeng Wang, Yaser Yacoob, and Lijuan Wang.
  2023{\natexlab{a}}.
\newblock Aligning large multi-modal model with robust instruction tuning.
\newblock \emph{arXiv preprint arXiv:2306.14565}.

\bibitem[{Liu et~al.(2023{\natexlab{b}})Liu, Sferrazza, and
  Abbeel}]{liu2023chain}
Hao Liu, Carmelo Sferrazza, and Pieter Abbeel. 2023{\natexlab{b}}.
\newblock Chain of hindsight aligns language models with feedback.
\newblock \emph{arXiv preprint arXiv:2302.02676}.

\bibitem[{Liu et~al.(2023{\natexlab{c}})Liu, Li, Wu, and
  Lee}]{DBLP:journals/corr/abs-2304-08485}
Haotian Liu, Chunyuan Li, Qingyang Wu, and Yong~Jae Lee. 2023{\natexlab{c}}.
\newblock Visual instruction tuning.
\newblock \emph{CoRR}.

\bibitem[{Lu et~al.(2019)Lu, Batra, Parikh, and Lee}]{DBLP:conf/nips/LuBPL19}
Jiasen Lu, Dhruv Batra, Devi Parikh, and Stefan Lee. 2019.
\newblock Vilbert: Pretraining task-agnostic visiolinguistic representations
  for vision-and-language tasks.
\newblock In \emph{Advances in Neural Information Processing Systems 32: Annual
  Conference on Neural Information Processing Systems 2019, NeurIPS 2019,
  December 8-14, 2019, Vancouver, BC, Canada}.

\bibitem[{Lu et~al.(2022{\natexlab{a}})Lu, Mishra, Xia, Qiu, Chang, Zhu,
  Tafjord, Clark, and Kalyan}]{DBLP:conf/nips/LuMX0CZTCK22}
Pan Lu, Swaroop Mishra, Tanglin Xia, Liang Qiu, Kai{-}Wei Chang, Song{-}Chun
  Zhu, Oyvind Tafjord, Peter Clark, and Ashwin Kalyan. 2022{\natexlab{a}}.
\newblock Learn to explain: Multimodal reasoning via thought chains for science
  question answering.
\newblock In \emph{NeurIPS}.

\bibitem[{Lu et~al.(2022{\natexlab{b}})Lu, Welleck, Hessel, Jiang, Qin, West,
  Ammanabrolu, and Choi}]{lu2022quark}
Ximing Lu, Sean Welleck, Jack Hessel, Liwei Jiang, Lianhui Qin, Peter West,
  Prithviraj Ammanabrolu, and Yejin Choi. 2022{\natexlab{b}}.
\newblock Quark: Controllable text generation with reinforced unlearning.
\newblock \emph{Advances in neural information processing systems}.

\bibitem[{Madaan and Yazdanbakhsh(2022)}]{DBLP:journals/corr/abs-2209-07686}
Aman Madaan and Amir Yazdanbakhsh. 2022.
\newblock Text and patterns: For effective chain of thought, it takes two to
  tango.
\newblock \emph{CoRR}.

\bibitem[{Meng et~al.(2022)Meng, Huang, Zhang, and
  Han}]{DBLP:conf/nips/MengHZH22}
Yu~Meng, Jiaxin Huang, Yu~Zhang, and Jiawei Han. 2022.
\newblock Generating training data with language models: Towards zero-shot
  language understanding.
\newblock In \emph{NeurIPS}.

\bibitem[{OpenAI(2023)}]{DBLP:journals/corr/abs-2303-08774}
OpenAI. 2023.
\newblock {GPT-4} technical report.
\newblock \emph{CoRR}.

\bibitem[{Ordonez et~al.(2011)Ordonez, Kulkarni, and
  Berg}]{DBLP:conf/nips/OrdonezKB11}
Vicente Ordonez, Girish Kulkarni, and Tamara~L. Berg. 2011.
\newblock Im2text: Describing images using 1 million captioned photographs.
\newblock In \emph{Advances in Neural Information Processing Systems 24: 25th
  Annual Conference on Neural Information Processing Systems 2011. Proceedings
  of a meeting held 12-14 December 2011, Granada, Spain}.

\bibitem[{Ouyang et~al.(2022)Ouyang, Wu, Jiang, Almeida, Wainwright, Mishkin,
  Zhang, Agarwal, Slama, Ray, Schulman, Hilton, Kelton, Miller, Simens, Askell,
  Welinder, Christiano, Leike, and Lowe}]{DBLP:conf/nips/Ouyang0JAWMZASR22}
Long Ouyang, Jeffrey Wu, Xu~Jiang, Diogo Almeida, Carroll~L. Wainwright, Pamela
  Mishkin, Chong Zhang, Sandhini Agarwal, Katarina Slama, Alex Ray, John
  Schulman, Jacob Hilton, Fraser Kelton, Luke Miller, Maddie Simens, Amanda
  Askell, Peter Welinder, Paul~F. Christiano, Jan Leike, and Ryan Lowe. 2022.
\newblock Training language models to follow instructions with human feedback.
\newblock In \emph{NeurIPS}.

\bibitem[{Poesia et~al.(2023)Poesia, Gandhi, Zelikman, and
  Goodman}]{DBLP:journals/corr/abs-2306-04031}
Gabriel Poesia, Kanishk Gandhi, Eric Zelikman, and Noah~D. Goodman. 2023.
\newblock Certified reasoning with language models.
\newblock \emph{CoRR}.

\bibitem[{Radford et~al.(2021)Radford, Kim, Hallacy, Ramesh, Goh, Agarwal,
  Sastry, Askell, Mishkin, Clark et~al.}]{radford2021learning}
Alec Radford, Jong~Wook Kim, Chris Hallacy, Aditya Ramesh, Gabriel Goh,
  Sandhini Agarwal, Girish Sastry, Amanda Askell, Pamela Mishkin, Jack Clark,
  et~al. 2021.
\newblock Learning transferable visual models from natural language
  supervision.
\newblock In \emph{International Conference on Machine Learning}.

\bibitem[{Ray et~al.(2019)Ray, Sikka, Divakaran, Lee, and
  Burachas}]{DBLP:conf/emnlp/RaySDLB19}
Arijit Ray, Karan Sikka, Ajay Divakaran, Stefan Lee, and Giedrius Burachas.
  2019.
\newblock Sunny and dark outside?! improving answer consistency in {VQA}
  through entailed question generation.
\newblock In \emph{Proceedings of the 2019 Conference on Empirical Methods in
  Natural Language Processing and the 9th International Joint Conference on
  Natural Language Processing, {EMNLP-IJCNLP} 2019, Hong Kong, China, November
  3-7, 2019}. Association for Computational Linguistics.

\bibitem[{Sahu et~al.(2022)Sahu, Cogswell, Gong, and
  Divakaran}]{DBLP:journals/corr/abs-2209-15093}
Pritish Sahu, Michael Cogswell, Yunye Gong, and Ajay Divakaran. 2022.
\newblock Unpacking large language models with conceptual consistency.
\newblock \emph{CoRR}.

\bibitem[{Sai et~al.(2023)Sai, Mohankumar, and
  Khapra}]{DBLP:journals/csur/SaiMK23}
Ananya~B. Sai, Akash~Kumar Mohankumar, and Mitesh~M. Khapra. 2023.
\newblock A survey of evaluation metrics used for {NLG} systems.
\newblock \emph{{ACM} Comput. Surv.}

\bibitem[{Saparov and He(2023)}]{DBLP:conf/iclr/Saparov023}
Abulhair Saparov and He~He. 2023.
\newblock Language models are greedy reasoners: {A} systematic formal analysis
  of chain-of-thought.
\newblock In \emph{The Eleventh International Conference on Learning
  Representations, {ICLR} 2023, Kigali, Rwanda, May 1-5, 2023}. OpenReview.net.

\bibitem[{Saparov et~al.(2023)Saparov, Pang, Padmakumar, Joshi, Kazemi, Kim,
  and He}]{DBLP:journals/corr/abs-2305-15269}
Abulhair Saparov, Richard~Yuanzhe Pang, Vishakh Padmakumar, Nitish Joshi,
  Seyed~Mehran Kazemi, Najoung Kim, and He~He. 2023.
\newblock Testing the general deductive reasoning capacity of large language
  models using {OOD} examples.
\newblock \emph{CoRR}.

\bibitem[{Selvaraju et~al.(2020)Selvaraju, Tendulkar, Parikh, Horvitz, Ribeiro,
  Nushi, and Kamar}]{DBLP:conf/cvpr/SelvarajuTPHRNK20}
Ramprasaath~R. Selvaraju, Purva Tendulkar, Devi Parikh, Eric Horvitz,
  Marco~T{\'{u}}lio Ribeiro, Besmira Nushi, and Ece Kamar. 2020.
\newblock Squinting at {VQA} models: Introspecting {VQA} models with
  sub-questions.
\newblock In \emph{2020 {IEEE/CVF} Conference on Computer Vision and Pattern
  Recognition, {CVPR} 2020, Seattle, WA, USA, June 13-19, 2020}. Computer
  Vision Foundation / {IEEE}.

\bibitem[{Tan and Bansal(2019)}]{tan2019lxmert}
Hao Tan and Mohit Bansal. 2019.
\newblock Lxmert: Learning cross-modality encoder representations from
  transformers.
\newblock In \emph{Proceedings of the 2019 Conference on Empirical Methods in
  Natural Language Processing}.

\bibitem[{Ubani et~al.(2023)Ubani, Polat, and
  Nielsen}]{DBLP:journals/corr/abs-2304-14334}
Solomon Ubani, Suleyman~Olcay Polat, and Rodney Nielsen. 2023.
\newblock Zeroshotdataaug: Generating and augmenting training data with
  chatgpt.
\newblock \emph{CoRR}.

\bibitem[{Uppal et~al.(2022)Uppal, Bhagat, Hazarika, Majumder, Poria,
  Zimmermann, and Zadeh}]{DBLP:journals/inffus/UppalBHMPZZ22}
Shagun Uppal, Sarthak Bhagat, Devamanyu Hazarika, Navonil Majumder, Soujanya
  Poria, Roger Zimmermann, and Amir Zadeh. 2022.
\newblock Multimodal research in vision and language: {A} review of current and
  emerging trends.
\newblock \emph{Inf. Fusion}.

\bibitem[{Walton(2014)}]{walton2014abductive}
Douglas Walton. 2014.
\newblock \emph{Abductive reasoning}.
\newblock University of Alabama Press.

\bibitem[{Wang et~al.(2022{\natexlab{a}})Wang, Min, Deng, Shen, Wu,
  Zettlemoyer, and Sun}]{DBLP:journals/corr/abs-2212-10001}
Boshi Wang, Sewon Min, Xiang Deng, Jiaming Shen, You Wu, Luke Zettlemoyer, and
  Huan Sun. 2022{\natexlab{a}}.
\newblock Towards understanding chain-of-thought prompting: An empirical study
  of what matters.
\newblock \emph{CoRR}.

\bibitem[{Wang et~al.(2022{\natexlab{b}})Wang, Yang, Men, Lin, Bai, Li, Ma,
  Zhou, Zhou, and Yang}]{DBLP:conf/icml/WangYMLBLMZZY22}
Peng Wang, An~Yang, Rui Men, Junyang Lin, Shuai Bai, Zhikang Li, Jianxin Ma,
  Chang Zhou, Jingren Zhou, and Hongxia Yang. 2022{\natexlab{b}}.
\newblock {OFA:} unifying architectures, tasks, and modalities through a simple
  sequence-to-sequence learning framework.
\newblock In \emph{International Conference on Machine Learning, {ICML} 2022,
  17-23 July 2022, Baltimore, Maryland, {USA}}. {Pmlr}.

\bibitem[{Wang et~al.(2022{\natexlab{c}})Wang, Chen, and
  Zhu}]{DBLP:journals/pami/WangCZ22}
Xin Wang, Yudong Chen, and Wenwu Zhu. 2022{\natexlab{c}}.
\newblock A survey on curriculum learning.
\newblock \emph{{IEEE} Trans. Pattern Anal. Mach. Intell.}

\bibitem[{Wang et~al.(2024)Wang, Chen, Yuan, Zhang, Li, Peng, and
  Ji}]{wang2024executable}
Xingyao Wang, Yangyi Chen, Lifan Yuan, Yizhe Zhang, Yunzhu Li, Hao Peng, and
  Heng Ji. 2024.
\newblock Executable code actions elicit better llm agents.
\newblock \emph{arXiv preprint arXiv:2402.01030}.

\bibitem[{Wang et~al.(2022{\natexlab{d}})Wang, Wei, Schuurmans, Le, Chi, and
  Zhou}]{DBLP:journals/corr/abs-2203-11171}
Xuezhi Wang, Jason Wei, Dale Schuurmans, Quoc~V. Le, Ed~H. Chi, and Denny Zhou.
  2022{\natexlab{d}}.
\newblock Self-consistency improves chain of thought reasoning in language
  models.
\newblock \emph{CoRR}.

\bibitem[{Wang et~al.(2022{\natexlab{e}})Wang, Kordi, Mishra, Liu, Smith,
  Khashabi, and Hajishirzi}]{DBLP:journals/corr/abs-2212-10560}
Yizhong Wang, Yeganeh Kordi, Swaroop Mishra, Alisa Liu, Noah~A. Smith, Daniel
  Khashabi, and Hannaneh Hajishirzi. 2022{\natexlab{e}}.
\newblock Self-instruct: Aligning language model with self generated
  instructions.
\newblock \emph{CoRR}.

\bibitem[{Wang et~al.(2022{\natexlab{f}})Wang, You, He, Li, Chang, and
  Chang}]{DBLP:conf/emnlp/WangYHLCC22}
Zhecan Wang, Haoxuan You, Yicheng He, Wenhao Li, Kai{-}Wei Chang, and Shih{-}Fu
  Chang. 2022{\natexlab{f}}.
\newblock Understanding me? multimodal evaluation for fine-grained visual
  commonsense.
\newblock In \emph{Proceedings of the 2022 Conference on Empirical Methods in
  Natural Language Processing, {EMNLP} 2022, Abu Dhabi, United Arab Emirates,
  December 7-11, 2022}. Association for Computational Linguistics.

\bibitem[{Wang et~al.(2021)Wang, Yu, Yu, Dai, Tsvetkov, and
  Cao}]{wang2021simvlm}
Zirui Wang, Jiahui Yu, Adams~Wei Yu, Zihang Dai, Yulia Tsvetkov, and Yuan Cao.
  2021.
\newblock Simvlm: Simple visual language model pretraining with weak
  supervision.
\newblock \emph{arXiv preprint arXiv:2108.10904}.

\bibitem[{Wei et~al.(2022)Wei, Wang, Schuurmans, Bosma, Ichter, Xia, Chi, Le,
  and Zhou}]{DBLP:conf/nips/Wei0SBIXCLZ22}
Jason Wei, Xuezhi Wang, Dale Schuurmans, Maarten Bosma, Brian Ichter, Fei Xia,
  Ed~H. Chi, Quoc~V. Le, and Denny Zhou. 2022.
\newblock Chain-of-thought prompting elicits reasoning in large language
  models.
\newblock In \emph{NeurIPS}.

\bibitem[{Wei et~al.(2023)Wei, Tan, Gao, Sun, Li, Yu, Guo, and
  Li}]{DBLP:journals/corr/abs-2307-12626}
Jingxuan Wei, Cheng Tan, Zhangyang Gao, Linzhuang Sun, Siyuan Li, Bihui Yu,
  Ruifeng Guo, and Stan~Z. Li. 2023.
\newblock Enhancing human-like multi-modal reasoning: {A} new challenging
  dataset and comprehensive framework.
\newblock \emph{CoRR}.

\bibitem[{Xu et~al.(2021)Xu, Yan, Li, Bi, Huang, Xiao, and
  Huang}]{DBLP:conf/acl/XuYLBHXH20}
Haiyang Xu, Ming Yan, Chenliang Li, Bin Bi, Songfang Huang, Wenming Xiao, and
  Fei Huang. 2021.
\newblock {E2E-VLP:} end-to-end vision-language pre-training enhanced by visual
  learning.
\newblock In \emph{Proceedings of the 59th Annual Meeting of the Association
  for Computational Linguistics and the 11th International Joint Conference on
  Natural Language Processing, {ACL/IJCNLP} 2021, (Volume 1: Long Papers),
  Virtual Event, August 1-6, 2021}. Association for Computational Linguistics.

\bibitem[{Xu et~al.(2023)Xu, Shao, Zhang, Gao, Liu, Lei, Meng, Huang, Qiao, and
  Luo}]{DBLP:journals/corr/abs-2306-09265}
Peng Xu, Wenqi Shao, Kaipeng Zhang, Peng Gao, Shuo Liu, Meng Lei, Fanqing Meng,
  Siyuan Huang, Yu~Qiao, and Ping Luo. 2023.
\newblock Lvlm-ehub: {A} comprehensive evaluation benchmark for large
  vision-language models.
\newblock \emph{CoRR}.

\bibitem[{Yang et~al.(2022)Yang, Gan, Wang, Hu, Ahmed, Liu, Lu, and
  Wang}]{DBLP:conf/eccv/YangGW000LW22}
Zhengyuan Yang, Zhe Gan, Jianfeng Wang, Xiaowei Hu, Faisal Ahmed, Zicheng Liu,
  Yumao Lu, and Lijuan Wang. 2022.
\newblock Unitab: Unifying text and box outputs for grounded vision-language
  modeling.
\newblock In \emph{Computer Vision - {ECCV} 2022 - 17th European Conference,
  Tel Aviv, Israel, October 23-27, 2022, Proceedings, Part {XXXVI}}. Springer.

\bibitem[{Yao et~al.(2023{\natexlab{a}})Yao, Yu, Zhao, Shafran, Griffiths, Cao,
  and Narasimhan}]{DBLP:journals/corr/abs-2305-10601}
Shunyu Yao, Dian Yu, Jeffrey Zhao, Izhak Shafran, Thomas~L. Griffiths, Yuan
  Cao, and Karthik Narasimhan. 2023{\natexlab{a}}.
\newblock Tree of thoughts: Deliberate problem solving with large language
  models.
\newblock \emph{CoRR}.

\bibitem[{Yao et~al.(2023{\natexlab{b}})Yao, Li, and
  Zhao}]{DBLP:journals/corr/abs-2305-16582}
Yao Yao, Zuchao Li, and Hai Zhao. 2023{\natexlab{b}}.
\newblock Beyond chain-of-thought, effective graph-of-thought reasoning in
  large language models.
\newblock \emph{CoRR}.

\bibitem[{Yao et~al.(2022)Yao, Chen, Zhang, Ji, Liu, Chua, and
  Sun}]{DBLP:conf/emnlp/YaoCZ0LCS22}
Yuan Yao, Qianyu Chen, Ao~Zhang, Wei Ji, Zhiyuan Liu, Tat{-}Seng Chua, and
  Maosong Sun. 2022.
\newblock {PEVL:} position-enhanced pre-training and prompt tuning for
  vision-language models.
\newblock In \emph{Proceedings of the 2022 Conference on Empirical Methods in
  Natural Language Processing, {EMNLP} 2022, Abu Dhabi, United Arab Emirates,
  December 7-11, 2022}. Association for Computational Linguistics.

\bibitem[{Yuan et~al.(2023)Yuan, Zhang, Chen, and
  Wei}]{DBLP:conf/acl/YuanZC023}
Lifan Yuan, Yichi Zhang, Yangyi Chen, and Wei Wei. 2023.
\newblock Bridge the gap between {CV} and nlp! {A} gradient-based textual
  adversarial attack framework.
\newblock In \emph{Findings of the Association for Computational Linguistics:
  {ACL} 2023, Toronto, Canada, July 9-14, 2023}. Association for Computational
  Linguistics.

\bibitem[{Zellers et~al.(2019)Zellers, Bisk, Farhadi, and
  Choi}]{zellers2019recognition}
Rowan Zellers, Yonatan Bisk, Ali Farhadi, and Yejin Choi. 2019.
\newblock From recognition to cognition: Visual commonsense reasoning.
\newblock In \emph{Proceedings of the IEEE/CVF Conference on Computer Vision
  and Pattern Recognition}.

\bibitem[{Zhang et~al.(2021)Zhang, Li, Hu, Yang, Zhang, Wang, Choi, and
  Gao}]{DBLP:conf/cvpr/ZhangLHY0WCG21}
Pengchuan Zhang, Xiujun Li, Xiaowei Hu, Jianwei Yang, Lei Zhang, Lijuan Wang,
  Yejin Choi, and Jianfeng Gao. 2021.
\newblock Vinvl: Revisiting visual representations in vision-language models.
\newblock In \emph{{IEEE} Conference on Computer Vision and Pattern
  Recognition, {CVPR} 2021, virtual, June 19-25, 2021}. Computer Vision
  Foundation / {IEEE}.

\bibitem[{Zhu et~al.(2023)Zhu, Chen, Shen, Li, and
  Elhoseiny}]{DBLP:journals/corr/abs-2304-10592}
Deyao Zhu, Jun Chen, Xiaoqian Shen, Xiang Li, and Mohamed Elhoseiny. 2023.
\newblock Minigpt-4: Enhancing vision-language understanding with advanced
  large language models.
\newblock \emph{CoRR}.

\end{thebibliography}

\newpage
\ 
\newpage
\newpage
\section*{Appendix}
\appendix

\begin{figure*}
  \begin{minipage}[b]{0.6\textwidth}
    \centering
    \includegraphics[width=\textwidth]{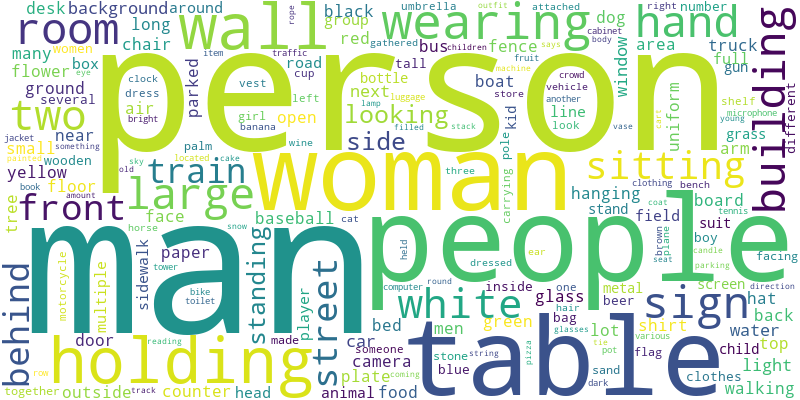} 
    \caption{The word cloud of the visual clues.}
    \label{fig:word_cloud}
  \end{minipage}
  \hfill
  \begin{minipage}[b]{0.35\textwidth}
    \centering
\begin{tabular}{c|c}
\toprule
Question Type & Percentage \\ \midrule
What          & 86.10      \\
Where         & 3.74       \\
Why           & 2.77       \\
How           & 2.16       \\
Which         & 1.84       \\
Who           & 1.54       \\
When          & 0.91       \\
Yes/No        & 0.68       \\
Others        & 0.25       \\ \bottomrule
\end{tabular}
\caption{Question distribution.}
    \label{tab:question_distribution}
  \end{minipage}
\end{figure*}


\section{Dataset Statistics}
\label{sec:stat}
\looseness=-1
\benchmark contains 1,622 evaluative instances, wherein each instance encompasses an average of 2.91 reasoning chains, also known as subquestions, reflecting a profound commitment to providing rich, complex data for effective analysis. 
On average, the lengths of the candidate inference, subquestions, and candidate answers in the dataset are 7.05, 9.97, and 2.96, respectively.
Note that these elements are products of LLMs, generated based on the visual clues provided by human annotators. 
We thus present the word cloud of the visual clues regarding the evaluation samples in Figure~\ref{fig:word_cloud}. 
Upon examination, it becomes apparent that these visual values primarily center around human-oriented concepts. They incorporate information about entities, activities, and occurrences that are directly associated with individuals.
This observation provides a partial representation of the data distribution within our dataset, particularly in relation to the target inference, subquestions, and their corresponding answers.

In addition, we delineate the distribution of question types within \benchmark as presented in Figure~\ref{tab:question_distribution}.
We find that \benchmark comprises various kinds of questions with the "What" type questions dominating the distribution.
This dominance is primarily due to the extensive use of such questions in Sherlock for cultivating a holistic comprehension of any given context or subject matter. Indeed, these types of queries are inherently employed to both obtain a detailed narrative of the scenario, as well as to facilitate visual inference based on the perceived information.
\section{Related Work}
\label{sec:related}
\paragraph{Vision-Language Pretraining.}
VLMs have demonstrated remarkable performance across various downstream tasks, primarily due to their extensive pre-training on large-scale datasets~\cite{DBLP:journals/ftcgv/GanLLWLG22, DBLP:journals/inffus/UppalBHMPZZ22, DBLP:journals/pami/WangCZ22}. 
Initially, VLMs heavily relied on object detectors for image comprehension~\cite{li2019visualbert, tan2019lxmert, DBLP:conf/nips/LuBPL19, DBLP:conf/aaai/LiDFGJ20, DBLP:conf/eccv/Li0LZHZWH0WCG20, DBLP:conf/naacl/LiYWZCC21, DBLP:conf/eccv/Li0LZHZWH0WCG20, DBLP:conf/naacl/LiYWZCC21, DBLP:conf/cvpr/ZhangLHY0WCG21}.
%
%
Subsequent developments in VLMs research have aimed to bypass the need for resource-intensive object detectors~\cite{DBLP:conf/cvpr/DouXGWWWZZYP0022, DBLP:journals/corr/abs-2004-00849, DBLP:conf/icml/KimSK21}, streamline the inference process~\cite{DBLP:conf/cvpr/HuangZH0FF21, DBLP:conf/acl/XuYLBHXH20}, incorporate more extensive visual data~\cite{DBLP:conf/eccv/YangGW000LW22, DBLP:conf/emnlp/YaoCZ0LCS22, li2021align, radford2021learning}, and introduce additional tasks for object grounding during pre-training~\cite{ DBLP:conf/icml/JiaYXCPPLSLD21, DBLP:conf/emnlp/YaoCZ0LCS22}.
As research progresses, efforts are made to design a unified architecture for VLMs, enabling them to handle multiple tasks without requiring task-specific adjustments~\cite{wang2021simvlm, DBLP:conf/icml/WangYMLBLMZZY22, DBLP:journals/corr/abs-2301-12597}.
Leveraging large-scale multimodal instruction tuning data for effective alignment of the two modalities, VLMs can effectively parse the questions and generate user-friendly responses~\cite{DBLP:journals/corr/abs-2301-12597, DBLP:journals/corr/abs-2304-08485, DBLP:journals/corr/abs-2304-10592}.

\paragraph{CoT Reasoning Consistency}
The CoT reasoning approach was initially introduced to enhance the reasoning capabilities of LLMs by prompting them to generate rationales and then answers~\cite{DBLP:conf/nips/Wei0SBIXCLZ22}. 
This approach is extended to various domains, models, and more complex problems~\cite{DBLP:journals/corr/abs-2306-04031, DBLP:conf/acl/LiHYRC023, DBLP:journals/corr/abs-2211-12588, DBLP:conf/acl/JinL23, DBLP:journals/corr/abs-2305-16582, DBLP:journals/corr/abs-2305-10601, DBLP:journals/corr/abs-2305-15269, wang2024executable}. 
In addition, the CoT reasoning consistency is effectively utilized to improve the reasoning performance~\cite{DBLP:journals/corr/abs-2203-11171}. 
However, it is still not clear how consistent LLMs reasoning is, given the mixed results in previous work~\cite{DBLP:journals/corr/abs-2212-10001, DBLP:journals/corr/abs-2307-13702, DBLP:journals/corr/abs-2209-07686, DBLP:conf/iclr/Saparov023, DBLP:journals/corr/abs-2209-15093}.

\section{Implementation Details of Evaluation}
\label{sec:implementation}
Given that none of the VLMs under consideration has been trained on grounded data, it is not feasible to directly incorporate bounding box information into these models
We adopt a compromise solution that involves preprocessing the evaluation samples through the automatic incorporation of annotated bounding boxes into the images. 
We instruct VLMs to focus on the specific region delineated by the bounding boxes in the prompts provided.
We describe the prompts for evaluation in Appendix~\ref{sec:prompt}.
For each top-tier question or subquestion in the reasoning chain, VLMs only need to select one option from candidate answers.
Namely, VLMs choose an answer based on the highest probability among six options: "A", "B", "C", "D", "E", and "F". 
 \begin{figure*}[tbp!]
\centering
\includegraphics[width=\textwidth]{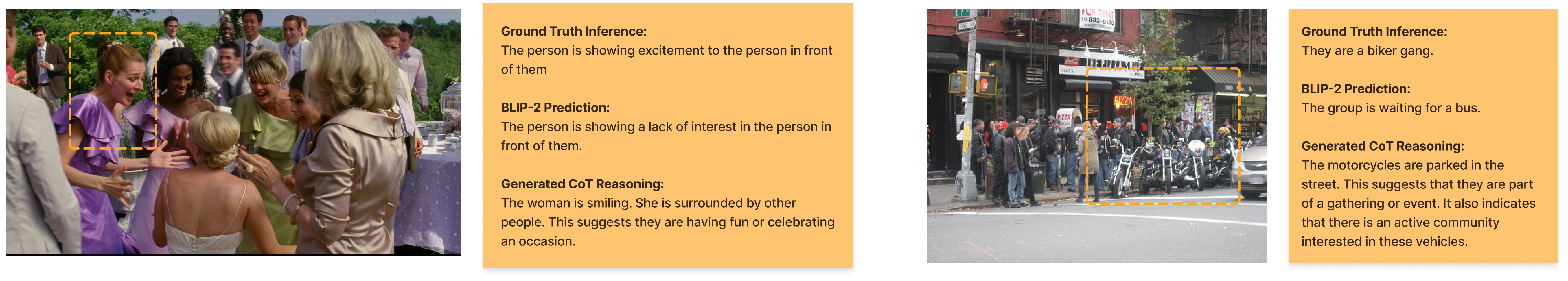}
 \caption{
 \looseness=-1
Showing cases where BLIP-2 fails initially but CoTBLIP is able to generate reasonable CoT reasoning chains that can help with the high-level visual inference and obtain the correct answer. More examples are in Figure~\ref{fig:more_examples_case}.
 } 
 \label{fig:case_study}
 \end{figure*}

\looseness=-1
\section{Human Annotation}
\label{sec:human_annotation}
\paragraph{Human Verification.}
We use the molardata platform\footnote{\url{https://www.molardata.com/}} for human annotation.
We hire 3 human annotators to validate each instance in the dataset generated by LLM, adhering to two specific criteria. 
First, the evaluation samples must be capable of effectively measuring reasoning performance and consistency. 
This entails instructing the annotators to examine the failure modes identified in Table~\ref{tab:filter} and also to identify any additional reasons for excluding certain examples from the evaluation;
Second, we plan to improve the diversity of the evaluation dataset by reducing instances that demonstrate highly similar reasoning paths within certain groups.
To this end, we provide each annotator with 100 dataset samples at the beginning of the annotation, acquainting them with the dataset's distribution as well as some analogous examples.
In the verification process,
we request annotators to label examples belonging to an extensive group of analogous cases.
Note that this is a dynamic process, as annotators have the capability to continuously update their understanding of the dataset distribution while engaging in the annotation task.
After completing the annotation process, we compile the results and subsequently exclude instances that have been classified as failures by any of the annotators.
We systematically collect examples labeled as relatively abundant in the dataset and subject them to a thorough validation process. 
We ensure the inclusion of a specific quantity of high-quality examples in each group, proportionate to the sample size within each group.

\paragraph{Dataset Evaluation.}
We hire a different set of 3 human annotators to conduct a cross-validation of the dataset derived from the human verification process, following the same verification procedure. 
Additionally, these annotators are requested to perform the task on \benchmark, including the high-level visual inference and CoT reasoning subquestions, thus capturing the human performance score.


\section{Forward Reasoning Consistency}
\label{sec:forward}
We choose the highest-performing models, specifically BLIP-2 and CoTBLIP, for conducting a qualitative analysis of their forward reasoning consistency.
We selected these models since they exhibit significant performance improvements compared to text-only models.
We select two examples, shown in Figure~\ref{fig:case_study}, to highlight cases where BLIP-2 demonstrates a lack of forward reasoning consistency and where CoTBLIP can potentially offer assistance.
We observe that CoTBLIP demonstrates the ability to generate coherent rationales, starting with visual elements that are highly relevant to the image, and subsequently advancing towards more sophisticated visual inference that significantly impacts the prediction. 
For example, the reasoning chain in the second example in Figure~\ref{fig:case_study} seems to first identify some motorcyclists that are parked on a street in some kind of gathering and then provides the high-level inference indicating that these folks might be part of a community interested in such vehicles. 
Notably, incorporating the rationales explicitly within the context enhances the reasoning consistency of VLMs.

\section{Prompts}
\label{sec:prompt}
We compile the list of prompts utilized in our implementation to instruct LLMs to perform their designated tasks.
\subsection{Candidate Answers Generation for CoT Subquestions}

\begin{lstlisting}
Imagine a scene: {Human-Annotated Visual Clue}
Here is a question and its corresponding answer:
Q: {CoT Subquestion}
A: {CoT Answer}
Please generate 5 different answers in 1-5 words that are semantically similar but contain factual errors.
\end{lstlisting}

\subsection{Filtering}
\begin{lstlisting}
    Imagine a scene: {Human-Annotated Visual Clue}
We want to make an inference about this scene by asking some coherent and subsequent questions. 
The questions are as follows:
Q1: {CoT Subquestion-1}
A1: {CoT Answer-1}
Q2: {CoT Subquestion-2}
A2: {CoT Answer-2}

So we draw the inference: {Human-Annotated High-Level Inference}

Judge whether the reasoning chain is coherent and consistent. Directly answer yes or no.
\end{lstlisting}

\subsection{Baseline Evaluation}
\begin{lstlisting}
    Pay attention to the designated area outlined by the red bounding box in the image.
Question: {Subquestion}
Six potential answers are as follows:
A: {Candidate A}
B: {Candidate B}
C: {Candidate C}
D: {Candidate D}
E: {Candidate E}
F: {Candidate F}
Which one is most likely to be correct? Directly answer (A/B/C/D/E/F):
\end{lstlisting}
\begin{lstlisting}
What can we infer from the designated area outlined by the red bounding box in the image? 
Here are six potential answers:
A: {Candidate A}
B: {Candidate B}
C: {Candidate C}
D: {Candidate D}
E: {Candidate E}
F: {Candidate F}
Which one is most likely to be correct? Directly answer (A/B/C/D/E/F):
\end{lstlisting}
\begin{lstlisting}
    What can we infer from the designated area outlined by the red bounding box in the image? Consider the following reasoning chain: {Reasoning Chain Generated by CoTBLIP}
Here are six potential answers:
A: {Candidate A}
B: {Candidate B}
C: {Candidate C}
D: {Candidate D}
E: {Candidate E}
F: {Candidate F}
Which one is most likely to be correct? Directly answer (A/B/C/D/E/F):
\end{lstlisting}

\subsection{CoT Reasoning Chains Generation}
\begin{lstlisting}
    You need to generate some questions for evaluating vision-language models. You will be given a scene description and a corresponding high-level inference about this scene.  Please generate step-by-step questions and corresponding answers that can derive the final inference. The reasoning chain should contain 2-4 questions, and the answers should contain 1-3 words.


Consider the following principle: 
1. The reasoning chain needs to be as short as possible.
2. The questions are used to evaluate vision-language models that don't have access to the scene description. So the first few questions are about visual information in the scene description, and you should not mention "description" in the questions.   
3. The reasoning chain should be consistent and cohesive. Each step should be atomic or based on previous steps, and should not be duplicated or redundant.
4. You should avoid generating questions with yes/no as the answers.
5. End your answer with the format 'The final reasoning chain is: ', and if you think such a task cannot be accomplished, please directly return 'No'. 


Scene description: patches of snow spread throughout grass on the side of freeway.
High-level inference: Cold weather is causing hazardous conditions at this location.

Let's think step by step: We need to initially generate some perceptual questions based on the visual information in the scene description. Then we need to generate questions about the visual reasoning based on the previously perceived information.  For the perceptual question, we have the information that patches of snow appear on the side of freeway in the scene. Then for the visual reasoning problem, we can infer that cold weather causes the appearance of snow, and based on the knowledge that snow can affect traffic, we can infer that cold weather can cause hazardous conditions at this location. 

The final reasoning chain is: 
Q1: What is seen on the grass on the side of the freeway?
A1: Patches of snow.
Q2: What kind of weather conditions could cause patches of snow to appear?
A2: Cold weather.
Q3: How can cold weather and patches of snow affect the conditions of a location?
A3: Hazardous conditions.


Scene description: {Human-Annotated Visual Clue}
High-level inference: {Human-Annotated High-Level Inference}

Let's think step by step:
\end{lstlisting}

\subsection{Postprocessing of LLaVA Dataset}
\begin{lstlisting}
    You need to generate some training samples for vision-language models. You will be given several scene descriptions to help you understand the image. Then a human-annotated question-answering pair will be given. 

Please generate a step-by-step reasoning chain. The reasoning chain should be very concise and short, containing less than 15 words for each step, and the total steps should be less than 4. 

For example: 
Scene description: 
A group of people skiing down a hill;
Several people on skis on a snowy slope;
A group of young men riding down the side of a snow covered slope;
Five skiers going through obstacles on a ski slope; 
people skating on the snow with orange flags on the way.

Question: What can you say about the skill level of the skiers featured in the image?
Answer: In the image, there is a group of five people skiing down a snowy slope with orange flags marking their trail. It appears that they are maneuvering through those obstacles on the ski slope, which suggests that these skiers possess a certain level of skill and experience. The fact that they can ski together in close proximity and navigate through obstacles demonstrates their ability to maintain control and balance while skiing. It's reasonable to assume that these skiers may have had some training or practice to reach this skill level, as navigating a snow-covered slope with obstacles typically requires a level of expertise beyond that of a beginner skier.

Let's think step by step:
The skiers are navigating through obstacles. They are skiing in close proximity. This implies advanced skill and experience.

Now consider this example:
Scene description: {COCO Captions}
Question: {LLaVA Question}
Answer: {LLaVA Answer}
Let's think step by step:
\end{lstlisting}

\subsection{Feedback Generation}
\begin{lstlisting}
    You will receive two inferences on an image produced by two models. Your objective is to choose the superior inference. Additionally, you will receive a description of the scene to aid in your understanding.

When making your judgment, please keep the following principles in mind:
1. The language used in the inference should flow naturally.
2. The inference should be grounded in the image's content.
3. The inference should be logical, congruent, and concise.
4. The inference should be comprehensive and complete, ultimately drawing a high-level inference.

Give your answer after the "Answer:"

 For example: 
Scene description: grass in a rock
Inferences:
A: The small plant growing out of the rock in the image is likely to be a native species that has been transplanted from another area, such as an urban or natural environment. Native plants tend to have deeper roots and are more resistant to environmental conditions than their non-native counterparts, making them ideal for adapting to different habitats and environments. 
B: The small plants growing out of the crack in the rock are likely a result of natural processes, such as erosion or weathering over time. This can be beneficial to the plant's survival and growth, helping it adapt to its environment and thrive. Additionally, this type of habitat is conducive for microorganisms that help maintain soil quality and provide nutrients necessary for healthy plant growth. 

Let's think step by step: Inference A captures the unique phenomenon of plants growing out of the rock, emphasizes the natural process and potential environmental benefits, and provides a more comprehensive understanding of the scene. On the contrary, Inference B fails to explain the mechanism or the specific relationship between the plants and the rock. The mention of adaptation to harsh conditions is relevant, but it does not encompass the entire context of the image. 
Answer: A

Now consider this example:
Scene description: {Image Caption}
Inference:
A: {Inference-1}
B: {Inference-2}

Let's think step by step:
\end{lstlisting}

%
%
%
%

%
%


 \begin{figure*}[t!]
\centering
\includegraphics[width=\textwidth]{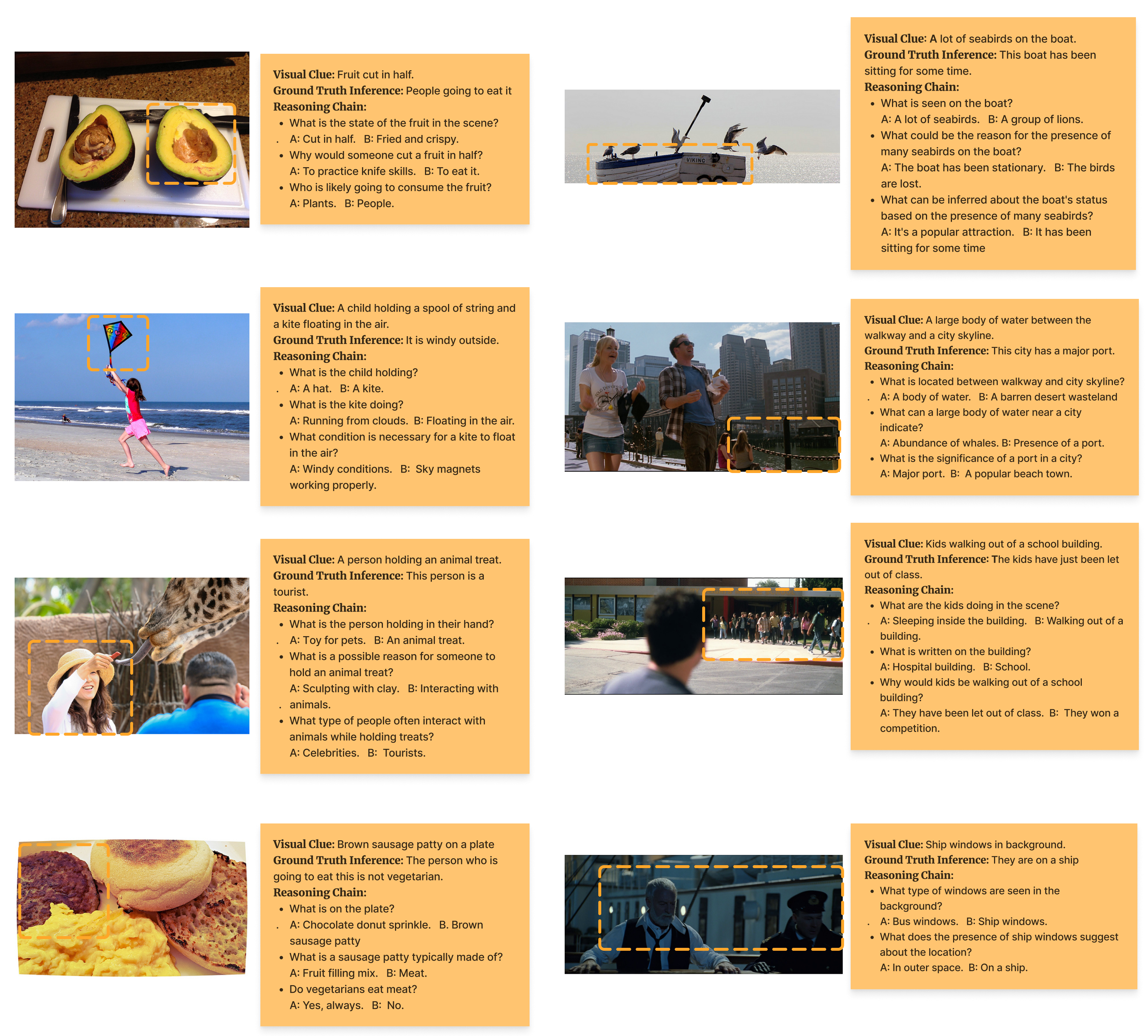}
 \caption{More examples included in \benchmark. We only show 2 candidate options (of 6 in total) for the sake of
presentation} 
 \label{fig:more_examples}
 \end{figure*}

 \begin{figure*}[t!]
\centering
\includegraphics[width=\textwidth]{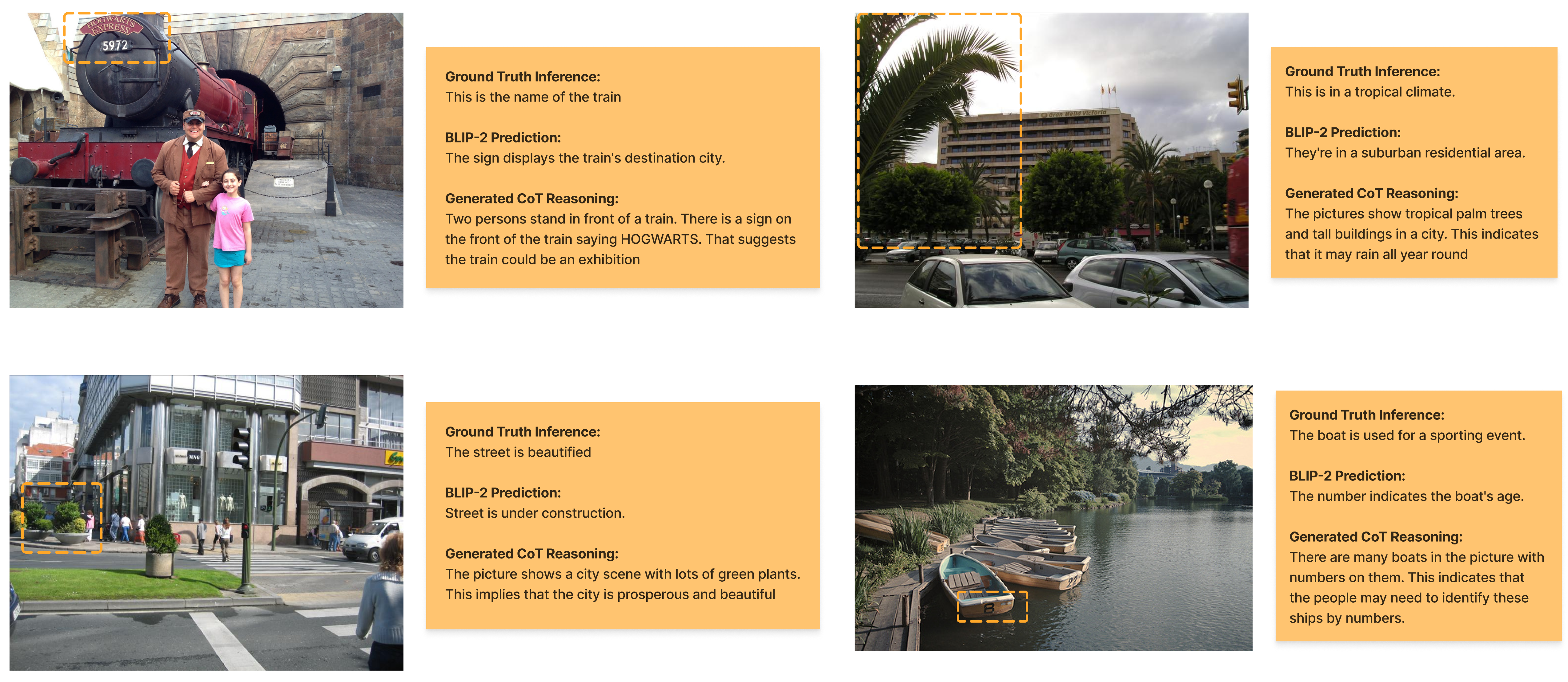}
 \caption{More examples for qualitative analysis of CoTBLIP.} 
 \label{fig:more_examples_case}
 \end{figure*}

\end{document}